\documentclass{article}

\usepackage{microtype}
\usepackage{graphicx}
\usepackage{subcaption}
\usepackage{booktabs}
\usepackage{adjustbox}
\usepackage{multirow}
\usepackage{array}
\usepackage{xcolor}
\usepackage{colortbl}
\usepackage{hyperref}
\usepackage{enumitem}
\usepackage[accepted]{icml2026}
\usepackage{amsmath,amssymb,mathtools,amsthm}
\usepackage[capitalize,noabbrev]{cleveref}      
\usepackage{graphicx}
\usepackage{soul} 

\definecolor{bestgreen} {HTML}{1B6E3F}
\definecolor{secondblue}{HTML}{1F4E9C}
\definecolor{textbg}    {HTML}{E1ECF7}  % soft sky blue (text-only intervention)
\definecolor{priorbg}   {HTML}{FCE2D2}  % soft peach   ($\pnum$ zeroed)
\definecolor{basebg}    {HTML}{F0F0F0}  % neutral grey (Original / reference)
\definecolor{flagbg}    {HTML}{FFF1B8}  % soft cream   (unimodal baseline)

\newcommand{\best}[1]{\textcolor{bestgreen}{\textbf{#1}}}
\newcommand{\second}[1]{\textcolor{secondblue}{\underline{#1}}}

\newcommand{\keystat}[1]{\textbf{#1}}

\newcommand{\TaTS}{\textsc{TaTS}}
\newcommand{\MMTSF}{\textsc{MM-TSFlib}}
\newcommand{\Aurora}{\textsc{Aurora}}
\newcommand{\TimeMMD}{Time-MMD}
\newcommand{\pname}{\texttt{prior\_history\_avg}}

\newcommand{\xnum}{\mathbf{x}}
\newcommand{\Ttext}{\mathbf{T}}
\newcommand{\pnum}{\mathbf{p}}
\newcommand{\yhat}{\hat{\mathbf{y}}}
\newcommand{\ytarget}{\mathbf{y}}
\newcommand{\Etext}{\mathbf{E}}
\newcommand{\Ztext}{\mathbf{Z}_{\!T}}
\newcommand{\Real}{\mathbb{R}}
\newcommand{\bbE}{\mathbb{E}}
\DeclareMathOperator{\MSE}{MSE}

\makeatletter
\renewcommand\paragraph{%
  \@startsection{paragraph}{4}{\z@}%
    {1.1ex \@plus 0.2ex \@minus 0.2ex}{-0.6em}%
    {\normalfont\normalsize\bfseries}}
\makeatother

\icmltitlerunning{Semantics or Structure? Auditing Text Sensitivity in Multimodal Time-Series Forecasting}

\begin{document}

\twocolumn[
  \icmltitle{Semantics or Structure?\\
             Auditing Text Sensitivity in
             Multimodal Time-Series Forecasting }
             
\icmlsetsymbol{super}{$\dagger$} 

\begin{icmlauthorlist} 
\icmlauthor{Karthik Sridhar}{bail} 
\icmlauthor{Atharva Gupta}{bits} 
\icmlauthor{Nishant Pradhan}{bits} 
\icmlauthor{Murari Mandal}{bail,kiit} 
\icmlauthor{Dhruv Kumar}{bail,bits,super} 
\icmlauthor{Saurabh Deshpande}{bail,super} 
\end{icmlauthorlist} \icmlaffiliation{bail}{Birla AI Labs, Mumbai, India} 
\icmlaffiliation{kiit}{KIIT, Bhubaneswar, India} 
\icmlaffiliation{bits}{BITS Pilani, Pilani, India} 
\icmlcorrespondingauthor{Saurabh Deshpande}{saurabh.deshpande-c@oab.adityabirla.com} \icmlcorrespondingauthor{Dhruv Kumar}{dhruv.kumar-c@oab.adityabirla.com}

  \icmlkeywords{multimodal time series, foundation models, probing,
                evaluation, structured data, TimeMMD}
  \vskip 0.3in
]
\printAffiliationsAndNotice{$\dagger$ Equal supervision. Code available at \url{github.com/birla-ai-labs/SemanticsOrStructure}.}
% ================================================================== %
\begin{abstract}
Multimodal time-series forecasting has emerged as a promising paradigm in which natural-language context is expected to improve predictive performance.
% Multimodal time-series forecasting is a promising paradigm in which
% natural-language text is expected to improve forecasting accuracy.
% The multimodal foundation model \Aurora{} and the late- and
% early-fusion paradigms \MMTSF{} and \TaTS{} all report significant
% improvements over unimodal baselines on the \TimeMMD{} benchmark,
% and attribute these gains to the text. Whether these models are
% sensitive to the \emph{content} of the text they receive has not
% been tested directly. 
Recent multimodal foundation models, including \Aurora{}, as well as early- and late-fusion approaches such as \MMTSF{} and \TaTS{}, report substantial gains over unimodal baselines on the \TimeMMD{} benchmark, attributing these improvements to textual information. However, whether these models are actually sensitive to the \emph{semantic content} of the text remains unverified. 
We address this question through controlled text perturbations, attribution analyses, and probes of \Aurora{}'s text pathway.
On \TimeMMD{}, swapping each row's text for any other
real text (empty, constant, within-domain shuffled, or cross-domain)
moves mean MSE by less than $0.5\%$ on all three architectures.
The improvement reported in the literature is recovered when a
co-shipped numeric column is removed without touching text. We
conclude that, on this benchmark and within this family of
frozen-encoder architectures, text content is not the operative
signal behind the reported gains. To support future work on text integration in multimodal foundation models for structured data, we release our perturbation protocol and evaluation harness as a reusable diagnostic toolkit.
%We hope these findings inform the design of future multimodal foundation models for structured data.
\end{abstract}

% ================================================================== %
\section{Introduction}
\label{sec:intro}

Many forecasting domains pair numeric series with temporally
aligned text. Epidemiological reports accompany case counts
\citep{wang2020covidml,liu2024timemmd}; energy dispatches accompany
consumption readings \citep{liu2024timemmd}; commodity bulletins and
financial news accompany price series \citep{ding2015deepevents,
araci2019finbert,sawhney2020stocknet}. The multimodal time-series
literature operationalises the intuition that this text encodes
regime changes, domain knowledge, and external events that numeric
histories cannot capture
\citep{jin2024timellm,jia2024gpt4mts,zhang2024llm4tssurvey,
       jin2024fmsurvey}. Three recent systems target this setting on
the \TimeMMD{}~\cite{liu2024timemmd} benchmark: the multimodal
\emph{foundation model} \Aurora{}~\cite{wu2026aurora}, and the
late- and early-fusion \emph{paradigms}
\MMTSF{}~\cite{liu2024timemmd} and \TaTS{}~\cite{li2025tats}. Each
provides a dedicated text pathway and reports significant MSE
improvements over its unimodal baseline.

The standard evaluation in this literature compares a full
multimodal run against a run with the text branch disabled. We
refer to the disabled-text configuration as the \emph{unimodal
baseline}, and to the MSE improvement of the multimodal run over
it as the \emph{multimodal lift}. This comparison reveals only that
the text pathway contributes \emph{some} signal. It does not reveal
whether the model is responding to what the text says: any fixed,
content-blind output of the text branch would produce the same
lift. The question of \emph{content sensitivity}, that is, whether
the model reacts to which words are in the input, has not been
studied systematically in multimodal time-series forecasting.
It parallels probing protocols in NLP that test whether models respond to
content or to spurious surface features
\citep{niven2019probing,mccoy2019hans}. Concurrent
work~\cite{zhang2025multimodality} asks whether multimodal gains
generalise \emph{across} datasets. We ask the orthogonal,
within-dataset question of whether the lift observed on \TimeMMD{}
is attributable to text content or to other inputs that share its
fusion path.

We test this by holding each of the three models fixed at its
released configuration and swapping the text input through five
substitutes while leaving every numeric input unchanged. Four of
these conditions probe content sensitivity to natural language
directly, by replacing the text with an empty string, a constant
placeholder, within-domain shuffled text, or text from a paired
domain. Together they span the natural-language hypothesis space.
A fifth \emph{oracle} condition replaces the text with a templated
sentence listing the future ground-truth values. We treat the
oracle as a necessary condition rather than a sufficient one: a
model that both responds to text and can read numeric prose must
register some change. We do not treat oracle insensitivity alone
as proof of content blindness, since the encoders these models use
may not represent numeric literals as quantities
\citep{wallace2019numbers}. None of the three architectures
responds meaningfully to any of the five substitutions.

\textbf{Contributions.}
\begin{itemize}[leftmargin=1.2em,topsep=2pt,itemsep=0pt]
\item A \textbf{text perturbation protocol} that varies text
  content independently of all other inputs, applied to three
  architectures on \TimeMMD{} spanning one foundation model and
  two paradigm implementations. All three are insensitive to
  natural-language text content within the precision of this
  benchmark.
\item \textbf{Mechanistic localisation} of the multimodal lift: a
  numeric column shipped with the dataset, blended into two
  architectures' outputs alongside the text, accounts for the gain.
  \Aurora{}-specific gradient, attention, and divergence probes
  show that its text pathway is trained but content-blind in the
  forward pass.
\end{itemize}
Code, patched runners, and the evaluation harness are released for
reproducibility.

% ================================================================== %
\section{Background}
\label{sec:background}

\paragraph{\TimeMMD{} as evaluation setting.}
\TimeMMD{}~\cite{liu2024timemmd} spans nine domains (Agriculture,
Climate, Economy, Energy, Environment, Health, Security, SocialGood,
Traffic). Each domain provides a numeric target series with
temporally aligned text drawn from public reports and domain
bulletins. Forecast horizons range from $6$ to $336$ steps depending
on sampling frequency. All three architectures we study were
originally evaluated on \TimeMMD{}, making it the natural testbed.

\paragraph{Three text pathways.}
Each architecture maps a numeric history $\xnum\in\Real^L$ and text
$\Ttext$ to a forecast $\yhat\in\Real^H$ through a model-specific
text pathway. Full architectural equations are in
Appendix~\ref{sec:app:arch}.
\textbf{\TaTS}~\cite{li2025tats} performs \emph{early fusion}. A
frozen GPT-2 pooler~\cite{radford2019gpt2} encodes each row's text
into a vector sequence. The sequence is projected and concatenated
to the numeric history along the channel axis, then passed to a
time-series backbone.
\textbf{\MMTSF}~\cite{liu2024timemmd} performs \emph{late fusion}.
A frozen BERT pooler~\cite{devlin2019bert} encodes the text into a
vector that is layer-normalised and added to the backbone's output
as a residual.
\textbf{\Aurora}~\cite{wu2026aurora} is a pretrained \emph{zero-shot}
foundation model. A frozen BERT-base produces per-token features
that a trained \emph{text distiller} compresses into $L_k\!=\!10$
learnable query tokens. A \emph{text guider} then injects them into
the temporal backbone via cross-attention.

All three architectures provide a dedicated text pathway. This
motivates the question of whether that pathway is sensitive to what
the text actually says.

\paragraph{The unimodal baseline and the multimodal lift.}
Each model ships a unimodal configuration that gates its text
pathway off, providing the natural reference for measuring
multimodal contribution. We define the \emph{multimodal lift} as
the MSE improvement of the full multimodal run over this unimodal
configuration. We reproduce the published claims in
Section~\ref{sec:results}.

% ================================================================== %
\section{Perturbation Protocol}
\label{sec:protocol}

We replace the text column with five substitutes and re-run each
model with all numeric inputs held identical across conditions.
Real \TimeMMD{} text carries a long preamble (e.g.\ ``\emph{Available
facts are as follows:\ldots}''); Table~\ref{tab:perturbations} shows
a schematic example for clarity.

\begin{table}[h]
\centering
\small
\setlength{\tabcolsep}{3pt}
\renewcommand{\arraystretch}{1.06}
\caption{\textbf{Five text conditions} on a schematic row from a
weather domain. Numeric history is $11.9$ in every condition.
The \textit{Tests} column names the property each substitution isolates.}
\label{tab:perturbations}
\begin{tabular}{@{}p{1.3cm}
                >{\raggedright\arraybackslash}p{4.15cm}
                >{\raggedright\arraybackslash}p{1.85cm}@{}}
\toprule
\textbf{Condition} & \textbf{Text fed to model} & \textbf{Tests} \\
\midrule
\textbf{Original}
  & ``Heavy rainfall expected tomorrow''
  & Baseline \\[2pt]
\textbf{Empty}
  & \textit{(empty string)}
  & Text presence \\[2pt]
\textbf{Constant}
  & ``Time series data point.''
  & Token content \\[2pt]
\textbf{Shuffled}
  & ``Hurricane expected'' \emph{(real text, different row, same domain)}
  & Row alignment \\[2pt]
\textbf{Cross-domain}
  & ``Trade balance deteriorated'' \emph{(date-aligned, paired domain)}
  & Topic relevance \\[2pt]
\textbf{Oracle}
  & ``Future values: $12.4, 12.8, 13.1$\ldots''
  & Numeric reading \\
\bottomrule
\end{tabular}
\end{table}

The first two conditions ablate the text entirely (no tokens, then
a fixed placeholder), testing whether the encoder reacts to mere
text \emph{presence}. The next two preserve real, fluent language
but sever its link to the row. \emph{Shuffled} keeps the text
distribution but breaks temporal alignment, while
\emph{Cross-domain} also breaks topical relevance. \emph{Oracle}
embeds the future itself. We treat it as a reference point rather
than a strict upper bound, since the encoders may not represent
numeric literals as quantities. Domain pairing details are in
Appendix~\ref{sec:app:perturbgen}. Each condition is compared to
the original-text baseline via the paired relative change
$\Delta_i\!=\!(\MSE_i(c)\!-\!\MSE_i(\text{orig})) / \MSE_i(\text{orig})$,
matching cells on (domain, horizon, seed, backbone). We aggregate
by the ratio of mean MSEs across cells, and attach $95\%$
bootstrap confidence intervals (CIs, $[2.5\%, 97.5\%]$
percentile)~\cite{efron1993bootstrap} with $10{,}000$ resamples
(Appendix~\ref{sec:app:cis}). We evaluate over
nine domains, four horizons, three seeds, and (for \TaTS{} and
\MMTSF{}) eight backbone variants.

% ================================================================== %
\section{Results}
\label{sec:results}

\begin{table}[h]
\centering
\small
\setlength{\tabcolsep}{3pt}
\renewcommand{\arraystretch}{1.08}
\caption{\textbf{Text perturbation results.} Mean test MSE averaged
over $9$ domains, $4$ horizons, $3$ seeds, and (for trained methods)
$8$ backbones. $\Delta\%$ is the change in mean MSE relative to the
original-text baseline. Bootstrap CIs and per-condition $p$-values
are in Appendix~\ref{sec:app:cis}, Table~\ref{tab:cis_full}.
\colorbox{textbg}{\phantom{x}} text-only perturbations;
\colorbox{flagbg}{\phantom{x}} unimodal baseline.
\best{Green bold}/\second{blue underline} mark the lowest/second-lowest
MSE per model. Plain $\mathbf{bold}$ flags structural $\Delta\%$ values
of magnitude $\geq 1\%$.}
\label{tab:main}
\begin{tabular}{@{}lcccccc@{}}
\toprule
 & \multicolumn{2}{c}{\Aurora} & \multicolumn{2}{c}{\MMTSF} & \multicolumn{2}{c}{\TaTS} \\
\cmidrule(lr){2-3}\cmidrule(lr){4-5}\cmidrule(lr){6-7}
\textbf{Condition} & MSE & $\Delta\%$ & MSE & $\Delta\%$ & MSE & $\Delta\%$ \\
\midrule
\rowcolor{basebg}
Original     & \second{8.553} & ---     & \second{14.03} & ---     & \best{13.19}   & ---     \\
\rowcolor{textbg}
Empty        & 8.555          & $+0.02$ & 14.04          & $+0.05$ & 13.19          & $-0.00$ \\
\rowcolor{textbg}
Constant     & 8.555          & $+0.03$ & 14.05          & $+0.16$ & 13.19          & $+0.00$ \\
\rowcolor{textbg}
Shuffled     & \second{8.553} & $+0.00$ & 14.03          & $-0.01$ & 13.19          & $+0.00$ \\
\rowcolor{textbg}
Cross-domain & \best{8.552}   & $-0.01$ & \best{14.03}   & $-0.03$ & 13.19          & $+0.00$ \\
\rowcolor{textbg}
Oracle       & 8.555          & $+0.02$ & 14.01          & $-0.14$ & 13.19          & $+0.00$ \\
\midrule
\rowcolor{flagbg}
Unimodal baseline & 8.553          & $+0.00$ & 14.29          & $\mathbf{+1.87}$ & 14.18          & $\mathbf{+7.54}$ \\
\bottomrule
\end{tabular}
\end{table}

\paragraph{Text content does not move forecasting error.}
Table~\ref{tab:main} delivers the headline finding. None of the
five text substitutions moves mean MSE by more than $0.5\%$ on any
of the three models. The largest shift is \keystat{$+0.16\%$} on
\MMTSF{} under Constant text. \TaTS{} stays within
\keystat{$\pm 0.001\%$}, and \Aurora{} stays within
\keystat{$\pm 0.05\%$}. This null pattern is consistent across
every condition and every model. Whether the text preserves natural
language (Shuffled, Cross-domain, Oracle), removes it (Empty), or
replaces it with a fixed placeholder (Constant), the forecast is
essentially unchanged. 

\paragraph{The unimodal lifts are real.}
The bottom row of Table~\ref{tab:main} confirms that enabling
the text pathway yields a positive multimodal lift, consistent
with the gains each method reports over its unimodal baseline.
What we cannot confirm is that this gain reflects the model
reading text content. If it did, substituting cross-domain text
or substituting text with future ground-truth values should
change something. It does not. (The oracle null in part reflects
that frozen GPT-2/BERT tokenise numerals as sub-word pieces that
do not preserve magnitude~\citep{wallace2019numbers}; the four
natural-language conditions carry the semantic weight.)

\paragraph{The null is robust across backbones.}
The mean numbers in Table~\ref{tab:main} aggregate over eight
backbones for \TaTS{} and \MMTSF{}. The multimodal lift varies
substantially with backbone choice: from \keystat{$+1.2\%$}
(iTransformer) to \keystat{$+15.0\%$} (Autoformer) on \TaTS{},
and from \keystat{$+0.3\%$} (FEDformer) to \keystat{$+3.8\%$}
(Autoformer) on \MMTSF{}. Yet text-only perturbations stay within
$\pm 0.01\%$ on \TaTS{} and $\pm 2.4\%$ on \MMTSF{} across all
backbones (Avg.\ column, Tables~\ref{tab:full_tats}, \ref{tab:full_mmtsf}). Backbone variance affects the
structural-column effect substantially; the text-content null
is uniform.

% ================================================================== %
\vspace{-2pt}
\section{Localising the Lift}
\label{sec:investigating}

The published multimodal lifts are real, yet text content does not
move error. Inspecting the architectures explains the gap.

\paragraph{A numeric column travels with the text.}
\TimeMMD{} ships alongside its text a numeric column $\pnum$
(\pname{}; Appendix~\ref{sec:app:datacolumns}) that stores a
numeric forecast derived from each row's target history. It is
a numeric feature, not a text feature. Both \TaTS{} and
\MMTSF{} blend $\pnum$ into the model output \emph{at the same
residual} as their text encoder.

In \TaTS{}, the final forecast is a convex combination of the
backbone output and $\pnum$ directly:
\begin{equation}
\yhat^{\TaTS}
  = (1-w)\,f_\theta\!\bigl([\xnum \,\|_{\!\mathrm{ch}}\, \psi(\Etext)]\bigr)
  + w\;\pnum_{L+1:L+H},
\label{eq:tats}
\end{equation}
where $f_\theta$ is the time-series backbone, $\psi(\Etext)$ is the
projected text embedding concatenated to $\xnum$ along the channel
axis, and $w\!=\!0.5$ by default, so half the forecast is literally
$\pnum$. In \MMTSF{}, the text embedding and $\pnum$ are
\emph{summed together} and then blended with the backbone:
\begin{equation}
\yhat^{\MMTSF}
  = (1-w)\,f_\theta(\xnum)
  + w\bigl(\mathrm{LN}(\bar\phi(\Ttext)) + \pnum_{L+1:L+H}\bigr),
\label{eq:mmtsflib}
\end{equation}
sharing a single residual gated by $w$. \Aurora{} never reads
$\pnum$ (Appendix~\ref{sec:app:arch}).

In both, the unimodal baseline gates text and $\pnum$ off
together through a single switch (Appendix~\ref{sec:app:arch}),
so the published lift reflects the \emph{combined} contribution
of text and $\pnum$. We note that \MMTSF{}'s appendix documents
$\pnum$ as a centering mechanism on the projection
output~\citep{liu2024timemmd}, while \TaTS{}'s framework
(Eq.~6--7 of \citet{li2025tats}) defines the forecast as
$F([\xnum;\,Z^{\top}])$ with no prior-mixing; the convex
combination in Eq.~\ref{eq:tats} appears in the released code,
not the paper.

\paragraph{A 2$\times$3 factorial isolates text from $\pnum$.}
Table~\ref{tab:prior} crosses a text axis (Original, Empty,
Constant) with a column axis (intact vs.\ zeroed). The
$2\!\times\!2$ block isolates the text effect from the $\pnum$
effect; the bottom row gives the unimodal baseline (both off).

\begin{table}[h]
\centering
\small
\setlength{\tabcolsep}{3.5pt}
\renewcommand{\arraystretch}{1.05}
\caption{\textbf{Disentangling text and $\pnum$ contributions.}
$\Delta\%$ MSE vs.\ the Original-text, $\pnum$-intact baseline,
with $95\%$ bootstrap CIs in brackets. Rows vary the text input;
columns toggle the numeric column $\pnum$. The contrast across
\emph{rows} measures the text effect; the contrast across
\emph{columns} measures the $\pnum$ effect.
\colorbox{textbg}{\phantom{x}} $\pnum$ intact;
\colorbox{priorbg}{\phantom{x}} $\pnum$ zeroed;
\colorbox{flagbg}{\phantom{x}} unimodal baseline.}
\label{tab:prior}
\begin{adjustbox}{max width=\columnwidth}
\begin{tabular}{@{}l ll ll ll@{}}
\toprule
& \multicolumn{2}{c}{\Aurora{}}
& \multicolumn{2}{c}{\MMTSF{}}
& \multicolumn{2}{c}{\TaTS{}} \\
\cmidrule(lr){2-3}\cmidrule(lr){4-5}\cmidrule(lr){6-7}
\textbf{Text}
& \cellcolor{textbg}\textbf{$\pnum$ in} & \cellcolor{priorbg}\textbf{$\pnum$=0}
& \cellcolor{textbg}\textbf{$\pnum$ in} & \cellcolor{priorbg}\textbf{$\pnum$=0}
& \cellcolor{textbg}\textbf{$\pnum$ in} & \cellcolor{priorbg}\textbf{$\pnum$=0} \\
\midrule
Original
 & \cellcolor{textbg}$0.00$  & \cellcolor{priorbg}$0.00$
 & \cellcolor{textbg}$0.00$  & \cellcolor{priorbg}$\mathbf{+1.52}$
 & \cellcolor{textbg}$0.00$  & \cellcolor{priorbg}$\mathbf{+20.16}$ \\
Empty
 & \cellcolor{textbg}$+0.02$ & \cellcolor{priorbg}$+0.02$
 & \cellcolor{textbg}$+0.05$ & \cellcolor{priorbg}$\mathbf{+1.62}$
 & \cellcolor{textbg}$-0.00$ & \cellcolor{priorbg}$\mathbf{+20.17}$ \\
Constant
 & \cellcolor{textbg}$+0.03$ & \cellcolor{priorbg}$+0.03$
 & \cellcolor{textbg}$+0.16$ & \cellcolor{priorbg}$\mathbf{+1.63}$
 & \cellcolor{textbg}$+0.00$ & \cellcolor{priorbg}$\mathbf{+20.16}$ \\
\midrule
\rowcolor{flagbg}
\textit{Unimodal}
 & \multicolumn{2}{c}{$+0.00$}
 & \multicolumn{2}{c}{$\mathbf{+1.87}$}
 & \multicolumn{2}{c}{$\mathbf{+7.55}$} \\
\bottomrule
\end{tabular}
\end{adjustbox}
\end{table}

\paragraph{Reading the factorial.}
\textbf{(i)}~Within any column (fixed $\pnum$ status), changing
the text moves MSE by less than $0.2\%$ on \MMTSF{}, $0.001\%$
on \TaTS{}, $0.05\%$ on \Aurora{}: text content is not the
operative signal.
\textbf{(ii)}~On \MMTSF{} ($w\!=\!0.1$), the $\pnum$-zeroed CI
$[+1.21,\!+1.85]$ overlaps the unimodal CI $[+1.50,\!+2.29]$;
the small blend weight perturbs amplitude by only ${\sim}10\%$,
so column-zeroing isolates $\pnum$'s contribution cleanly. The
published \MMTSF{} lift is accounted for by $\pnum$, with the
text residual indistinguishable from zero.
\textbf{(iii)}~On \TaTS{} ($w\!=\!0.5$), zeroing $\pnum$ halves
the prediction scale, so the $+20\%$ column-zeroed effect is
dominated by an amplitude artifact; the amplitude-matched
comparison is the unimodal baseline ($+7.5\%$). The text-content
null on \TaTS{} rests on the text-only perturbations, which stay
within $\pm 0.001\%$ even with the projection MLP made trainable
(Appendix~\ref{sec:app:detach}).

\paragraph{Backbone choice can hide the picture.}
On \TaTS{}, the $\pnum$-zeroed $\Delta\%$ varies from
\keystat{$+4.5\%$} (Autoformer) to \keystat{$+49.9\%$} (FiLM); on
\MMTSF{} the unimodal-baseline gap ranges from $+0.3\%$ to
$+3.8\%$ (Appendix~\ref{sec:app:fullresults},
Tables~\ref{tab:full_tats}, \ref{tab:full_mmtsf}). A lift on
a single backbone can over- or understate $\pnum$'s contribution
by an order of magnitude.

\paragraph{Aurora's text pathway is active but content-blind.}
Since \Aurora{} does not consume $\pnum$, its null result needs a
separate explanation. Three probes at the text-distiller interface
(Appendix~\ref{sec:app:probes}) ask whether the pathway was trained,
whether it discriminates between inputs, and whether it moves the
forecast.

\begin{table}[h]
\centering
\small
\setlength{\tabcolsep}{3pt}
\renewcommand{\arraystretch}{0.95}
\label{tab:probes_main}
\begin{adjustbox}{max width=\columnwidth}
\begin{tabular}{@{}l ccc@{}}
\toprule
& \multicolumn{1}{c}{\textit{Trained?}}
& \multicolumn{1}{c}{\textit{Discriminates?}}
& \multicolumn{1}{c}{\textit{Affects forecast?}} \\
\textbf{Condition}
& \textbf{Grad.\ norm} & \textbf{Attn.\ entropy} & \textbf{Pred.\ change} \\
& \textit{(non-zero}        & \textit{(low}          & \textit{(large} \\
& \textit{= trained)}       & \textit{= focused)}    & \textit{= text matters)} \\
\midrule
Original     & 0.15 & 0.975 & 0.041 \\
Empty        & 0.07 & 0.975 & 0.040 \\
Constant     & 0.16 & 0.976 & 0.044 \\
Shuffled     & 0.15 & 0.975 & 0.035 \\
Cross-domain & 0.07 & 0.975 & 0.035 \\
Oracle       & 0.13 & 0.976 & 0.058 \\
\bottomrule
\end{tabular}
\end{adjustbox}
\end{table}

Gradient norms are non-zero ($0.07$--$0.16$), so the pathway
\emph{was} optimised. Attention entropy sits at \keystat{$0.975$} in
every condition including oracle: nearly uniform across distilled
tokens regardless of text. Prediction change is at most
\keystat{$0.058$}, two orders of magnitude below the test-MSE scale
of $\sim 8.6$. The pathway is trained but content-blind in the
forward pass.

\paragraph{Is there a usable signal in \TimeMMD{} text?}
Having established content insensitivity across all three
architectures, we ask whether \TimeMMD{} text carries signal
that an attentive encoder \emph{could} exploit. We measure three
structural properties of the per-row text embeddings across all nine
domains: TTW (text-target Wasserstein distance, lower = better
alignment)~\cite{li2025tats}, ETA (embedding temporal
autocorrelation, high = persistent), and SDI (semantic diversity
index, high = distinct rows). Formal definitions are in
Appendix~\ref{sec:app:diagnostics}.

\begin{table}[h]
\centering
\small
\setlength{\tabcolsep}{6pt}
\renewcommand{\arraystretch}{1.0}
\caption{\textbf{Text diagnostics on \TimeMMD{}} (mean over $8$
domains, \textsc{Environment} excluded as self-paired). Per-domain
values and per-perturbation values are in
Appendix~\ref{sec:app:diagnostics},
Tables~\ref{tab:diag_full}--\ref{tab:diag_perturb}.}
\label{tab:diag_main}
\begin{tabular}{@{}lccc@{}}
\toprule
\textbf{Encoder} & \textbf{TTW} & \textbf{ETA} & \textbf{SDI} \\
& \textit{(low = aligned)} & \textit{(high = persistent)}
& \textit{(high = distinct)} \\
\midrule
GPT-2 & 0.056 & 0.423 & 0.008 \\
BERT  & 0.037 & 0.385 & 0.031 \\
\bottomrule
\end{tabular}
\end{table}

Table~\ref{tab:diag_main} shows moderate TTW and ETA but uniformly
low SDI, so consecutive rows produce nearly identical embeddings.
Our perturbations move all three diagnostics substantially yet
downstream MSE still stays within $0.5\%$, so none of TTW, ETA, or
SDI reliably predicts whether a perturbation will move the forecast.

% ================================================================== %
\vspace{-10pt}
\section{Discussion and Conclusion}
\label{sec:conclusion}

On \TimeMMD{}, substituting any plausible text for the original
(empty, constant, shuffled, cross-domain, or oracle) moves mean MSE
by less than $0.5\%$ on all three architectures; the lifts are real
but survive any text substitution. On \TaTS{} and \MMTSF{} they
localise to a numeric column ($\pnum$) co-routed through the same
fusion residual as the text encoder; on \Aurora{} the text pathway
is trained but content-blind in the forward pass.

The standard comparison against a disabled-text baseline cannot
distinguish these scenarios: a model that reads text and a model
that gates on a co-routed numeric prior both produce the same lift.
Established benchmarks for multimodal time-series forecasting
therefore do not verify that reported gains reflect genuine use of
text semantics; they verify only that the text pathway contributes
\emph{some} signal, not what kind. Rigorous multimodal benchmarking
requires direct text-content intervention and isolation of any numeric
features co-shipped through the text-fusion path. Richer corpora with higher per-row semantic
diversity would give attentive encoders something genuine to read.

\paragraph{Limitations.}
The audit covers \TimeMMD{} and three frozen-encoder architectures;
end-to-end trained encoders and higher-diversity benchmarks are
natural next steps. A directional oracle such as \emph{``a sharp
rise is expected''} would probe semantic sensitivity more directly
than our numeric oracle, which is bottlenecked by sub-word
tokenisation. We release our harness so this check becomes a default: what looks
like a text gain may be something else entirely.

% ================================================================== %
\section*{Impact Statement}

This paper presents an audit of multimodal time-series forecasting
methods on a public benchmark. The work is methodological: it does
not introduce new models, datasets, or applications, and uses only
the existing \TimeMMD{} dataset along with publicly released
research code. By identifying confounded baselines in published
results, the work aims to improve the rigour of evaluation
standards in this subfield. We see no specific ethical concerns,
direct deployment risks, or applications involving human subjects,
personally identifiable information, copyrighted training data, or
dual-use technologies that warrant further discussion. Beyond
contributing to the general advancement of machine learning, we do
not anticipate societal consequences specific to this work that
require highlighting.

% ================================================================== %
\bibliographystyle{icml2026}
\bibliography{references}

\appendix
% ================================================================== %
%  APPENDIX
% ================================================================== %

% Friendly aliases used in this appendix only
\newcommand{\Coriginal}{Original}
\newcommand{\Cempty}{Empty}
\newcommand{\Cconst}{Constant}
\newcommand{\Cshuf}{Shuffled}
\newcommand{\Ccross}{Cross-domain}
\newcommand{\Coracle}{Oracle}
\newcommand{\Czero}{Col.~zeroed}
\newcommand{\Cflag}{Unimodal}

% ================================================================== %
\section{Architecture Equations}
\label{sec:app:arch}

This section gives the full forward-pass equations for each
architecture and states formally how their unimodal baselines are
defined.

\paragraph{Notation.}
A \TimeMMD{} window of history length $L$ and horizon $H$ provides a
numeric history $\xnum\!\in\!\Real^L$, per-row text $\Ttext$, the
numeric column $\pnum\!\in\!\Real^{L+H}$ documented as \pname{}
(see Appendix~\ref{sec:app:datacolumns}), and target
$\ytarget\!\in\!\Real^H$. A frozen text encoder $\phi$ maps each
row's text to a vector; $\Etext(\Ttext)$ stacks these row-wise.

\paragraph{\TaTS{} (early fusion).}
A frozen GPT-2 pooler produces text features. A trainable
projection $\psi:\Real^{d_e}\!\to\!\Real^{d_p}$ is applied row-wise,
and the result is concatenated to $\xnum$ along the channel axis
before the backbone $f_\theta$. The final forecast is a convex
combination of the backbone output and the column $\pnum$:
\[
\yhat^{\TaTS}
  = (1-w)\,f_\theta\!\bigl([\xnum \,\|_{\!\mathrm{ch}}\, \psi(\Etext)]\bigr)
  + w\;\pnum_{L+1:L+H}.
\]
The default blend weight is $w\!=\!0.5$. The unimodal baseline is
recovered by setting both the projected text channels to zero
(text branch off) and $w\!=\!0$ ($\pnum$ contribution removed). A
single ablation thus removes both signals simultaneously, which is
why the unimodal lift in Table~\ref{tab:main} cannot be attributed
to the text channel alone.

\paragraph{\MMTSF{} (late fusion).}
A frozen BERT pooler produces a pooled text embedding, which is
projected to a horizon-shaped vector $\bar\phi(\Ttext)$,
layer-normalised, summed with $\pnum$, and blended with the
backbone:
\[
\yhat^{\MMTSF}
  = (1-w)\,f_\theta(\xnum)
  + w\bigl(\mathrm{LN}(\bar\phi(\Ttext)) + \pnum_{L+1:L+H}\bigr).
\]
The default blend weight is $w\!=\!0.1$. The unimodal baseline is
recovered by setting $w\!=\!0$. Because the text term and $\pnum$
share a single residual gated by $w$, this ablation also removes
both signals simultaneously.

\paragraph{\Aurora{} (cross-attention, zero-shot).}
A frozen BERT-base produces per-token features
$\Etext\!\in\!\Real^{L'\times d_e}$ ($L'\!=\!125$). A trained text
distiller compresses these into $L_k\!=\!10$ learnable query tokens
$Q$ via a transformer decoder. A text guider then injects them into
the temporal backbone via cross-attention:
\[
\Ztext = \text{TextDistill}(Q, \Etext),\quad
\alpha_t = \text{TextGuider}(\xnum, \Ztext),
\]
\[
\yhat^{\Aurora} = f^{\Aurora}_\theta(\xnum, \alpha_t, \Etext).
\]
\Aurora{} does not consume $\pnum$; it has no late-fusion residual
that would mix text and numeric features. The unimodal baseline is
the no-text variant of \Aurora{}; we expose a flag in our patched
runner that disables the text path during inference.

\paragraph{Summary of unimodal-baseline mechanics.}
On \TaTS{} and \MMTSF{}, the standard ``unimodal'' setting removes
both text and $\pnum$ together, because both pass through the same
gating coefficient. The published unimodal lift therefore reflects
the combined contribution of the text pathway and $\pnum$, not the
text alone. \Aurora{}'s unimodal baseline removes only text (it has
no $\pnum$ pathway), which is the cleanest of the three.

% ================================================================== %
\section{The Time-MMD Data Columns}
\label{sec:app:datacolumns}

The \TimeMMD{} benchmark ships per-domain CSV files containing
several columns whose use across architectures is not documented in
one place in the upstream literature. We reconstruct it here from
the data loaders of each repository.

\paragraph{Columns shared by all architectures.}
\textbf{\texttt{date}}: timestamp index. \textbf{\texttt{OT}}: the
numeric target series ($\ytarget$). \textbf{\pname{}}: a numeric
column that stores, for each row, an LLM-derived numeric forecast
of the target derived from the row's own preceding history. Despite
its name, this column is functionally close to a smoothed running
average of past target values; an LLM was prompted with the OT
history and asked to emit a numeric prediction. The output
correlates strongly with rolling-mean baselines but is not
identical to them. We refer to this column as $\pnum$ throughout
the paper.

\paragraph{MMTSFlib-specific columns.}
The MMTSFlib data files additionally contain text and feature
columns that originate from the Time-MMD construction pipeline.
\textbf{\texttt{Final\_Search\_2}}, \textbf{\texttt{Final\_Search\_4}},
\textbf{\texttt{Final\_Search\_6}}: the dataset was built by
retrieving relevant news articles or domain reports for each row's
date via web search; \texttt{Final\_Search\_N} concatenates $N$
retrieved chunks. The reference setting of MMTSFlib reads
\texttt{Final\_Search\_4} as its text column (controlled by
\texttt{--text\_len 4}). \textbf{\texttt{Final\_Output}}: a
closed-source LLM (likely GPT-4) processed the search results and
emitted a cleaned synthesis. MMTSFlib reads this column when the
\texttt{--use\_closedllm} flag is on; we use the default off
setting, which means the text input to the model is
\texttt{Final\_Search\_4}. \textbf{\texttt{his\_avg\_1..7}},
\textbf{\texttt{his\_std\_1..7}}: windowed averages and standard
deviations of OT at lookbacks of $1$ to $7$ periods. These columns
are present in the CSV but are \emph{not} consumed by MMTSFlib
under our evaluation setting (\texttt{features='S'}, see below).

\paragraph{TaTS-specific columns.}
\textbf{\texttt{fact}}: the text column TaTS actually uses, drawn
from the same source documents as MMTSFlib's
\texttt{Final\_Search\_4} but preprocessed slightly differently.
\textbf{\texttt{preds}}: an LLM-generated forward-looking
prediction for the target rendered as natural-language prose
(e.g.\ \emph{``the predicted value for next period is
2.34''}). This column exists in the CSV but TaTS's data loader
does not read it: only \texttt{fact} is bound to the
\texttt{self.text} attribute. The \texttt{preds} column was likely
pre-generated for a different model variant or for analysis. We
note that \texttt{preds} would have provided a softer
oracle-style condition (LLM-predicted future values, rather than
ground truth) but is not part of our evaluation, since TaTS's
released code does not consume it.

\paragraph{Aurora data.}
\Aurora{} reuses the TaTS CSV files. Its data loader reads
\texttt{fact} as the text input and ignores all other text columns.
\Aurora{} does not consume \pname{}.

\paragraph{features='S' versus features='M'.}
The data loaders inherited from the upstream time-series-library
support a \texttt{features} flag with two relevant settings.
\textbf{Univariate (\texttt{features='S'})}, which we use throughout,
reads only OT as the time-series input, with text handled on a
separate code path and \pname{} entering as the residual term in
the fusion equations above.
\textbf{Multivariate (\texttt{features='M'})} would attempt to read
all non-date columns into the time-series input matrix. This setting
crashes on the shipped CSVs across all three architectures, because
the inherited loader treats text columns
(\texttt{Final\_Search\_4}, \texttt{fact}) as numeric features and
StandardScaler raises a type error on the string contents. The
upstream papers' \texttt{features='M'} runs were performed against
pre-processed numeric-only CSVs that are not shipped with the
public benchmark. Switching to \texttt{features='M'} would also
introduce \texttt{his\_avg\_1..7} and \texttt{his\_std\_1..7} as
backbone inputs, which would change the experiment in a non-trivial
way. We retain \texttt{features='S'} for cleanliness and
consistency across all three models.

% ================================================================== %
\section{Methodology}
\label{sec:app:method}

\paragraph{Repository preparation.}
We pin each upstream repository to a specific commit and apply a
small set of documented, idempotent patches falling into five
categories.
\textbf{(a)}~CSV-loading fixes that prevent silent conversion of
empty text strings to \texttt{NaN}, which would otherwise corrupt
the empty-text condition by replacing the empty string with a
literal \texttt{"nan"} or with sentinel text.
\textbf{(b)}~Pandas-version compatibility fixes.
\textbf{(c)}~Backbone registration in \TaTS{}, which originally
registered only iTransformer in its model registry; we register the
seven additional backbones (Autoformer, Crossformer, DLinear,
FEDformer, FiLM, Informer, Transformer) already present in its
source tree.
\textbf{(d)}~An \Aurora{} command-line flag for the unimodal
ablation, since \Aurora{}'s released code does not expose one.
\textbf{(e)}~A \TaTS{} \texttt{--fix\_text\_grad} flag that
restores gradient flow into the trainable text-projection MLP
(see Appendix~\ref{sec:app:detach} for the full description). All
\TaTS{} results in this paper use this patch.

\paragraph{Perturbation generation.}
We produce one perturbed CSV per (condition, seed, domain). Row
count, the date column, and the numeric target column are preserved
exactly; a post-hoc validator confirms every perturbed file matches
the original on all unchanged columns. Construction details for
each condition appear in Appendix~\ref{sec:app:perturbgen}.

\paragraph{Per-cell evaluation.}
A run is a single (model, backbone, condition, seed, domain,
horizon) tuple. \Aurora{} is evaluated zero-shot: pretrained
weights are loaded once and used for inference, with the seed
controlling only the flow-matching head's stochastic sampling (we
average $100$ samples per cell, the value used in the released
script). Per-domain sequence lengths follow \Aurora{}'s reference
defaults (e.g.\ $L\!=\!192$ for Agriculture, $L\!=\!1056$ for
Energy). \TaTS{} and \MMTSF{} are fine-tuned per cell on the
perturbed training split for five epochs with patience five (each
repository's own defaults), with univariate target features. The
backbone, text encoder, and blend weight follow each model's
reference defaults; only the perturbed CSV varies between
conditions.

\paragraph{Determinism.}
Each runner sets the PyTorch and NumPy seeds per cell, and pins
GPU visibility per shard. We do not enforce strict deterministic
algorithms because one \MMTSF{} attention path is incompatible with
that mode. Run-to-run reproducibility on a fixed seed is bit-exact
for \Aurora{} and within floating-point non-associativity (around
$13$ significant figures) for the trained methods.

\paragraph{Probes.}
After fitting, we reload each \Aurora{} cell, attach forward and
backward hooks at the distilled-token interface, and compute the
three quantities defined in Appendix~\ref{sec:app:probes}. We
verified that attaching the hooks does not perturb the forward
graph: forward outputs match bit-exactly with and without the
hooks.

% ================================================================== %
\section{Perturbation Generation Details}
\label{sec:app:perturbgen}

\paragraph{Empty.}
The text column is set to the empty string for every row.

\paragraph{Constant.}
The text column is set to ``Time series data point.'' for every
row, a non-empty token sequence with no row-specific content.

\paragraph{Shuffled.}
The text column is permuted within each domain via a permutation
seeded by the run seed. All text fields belonging to a row are
permuted with the same permutation, preserving cross-column
alignment within the row. This is the only condition whose CSV
content depends on the run seed.

\paragraph{Cross-domain.}
We use a fixed pairing:
Agriculture\,$\leftrightarrow$\,Security,
Climate\,$\leftrightarrow$\,Energy,
Economy\,$\leftrightarrow$\,Health,
SocialGood\,$\leftrightarrow$\,Traffic;
Environment is self-paired and falls back to within-domain shuffle.
For each target row we use the paired domain's row whose date is the
latest available date not exceeding the target row's date, with
deterministic tie-breaking. This makes the condition seed-independent
and avoids using information from the target row's future.

\paragraph{Oracle.}
For each row in the train, validation, and test splits we
substitute a templated string of the form ``\emph{Available facts
are as follows: Step+1: The target will be $y_1$. Step+2: The
target will be $y_2$.\ldots}'' using the row's ground-truth future
target values. The substitution is applied to the train and
validation splits as well, so the fine-tuned methods see consistent
oracle structure during training.

\paragraph{Column-zeroed conditions.}
The numeric column $\pnum$ is set to zero for every row; the text
column varies (intact, empty, or constant). \Aurora{} does not read
$\pnum$, so on \Aurora{} these conditions are equivalent to the
corresponding text-only substitutions, and we use them as an
internal control.

% ================================================================== %
\section{Probe Definitions and Full Results}
\label{sec:app:probes}

The three probes target the interface between \Aurora{}'s frozen
BERT encoder and its trainable text distiller, where the model
collapses the per-token features into the $L_k\!=\!10$ distilled
tokens that the temporal backbone reads via cross-attention. Each
probe asks a different question.

\paragraph{Probe A: gradient norm at the distilled tokens.}
Given the distilled-token tensor $\Ztext$, we measure the
root-mean-square loss-gradient flowing back through it,
\[
\widehat{g}
= \sqrt{\tfrac{1}{B L_k d}\,\textstyle\sum_{b,k,j}
        \bigl[\nabla_{\Ztext}\mathcal{L}\bigr]_{b,k,j}^{2}}.
\]
This asks whether the trainable text path receives non-trivial
gradient signal during training. A value of $\widehat{g}\!=\!0$
would mean the path was never optimised.

\paragraph{Probe B: normalised cross-attention entropy.}
For guider weights $\alpha_t$ over the $L_k\!=\!10$ distilled
tokens,
\[
\widehat{H}
= \tfrac{1}{\log L_k}\,
  \bbE_{b,h,q}\!\Bigl[-\textstyle\sum_{k}(\alpha_t)_{bhqk}
                         \log(\alpha_t)_{bhqk}\Bigr].
\]
$\widehat{H}\!=\!1$ corresponds to perfectly uniform attention and
$\widehat{H}\!=\!0$ to a one-hot focus on a single token. This asks
whether the guider \emph{discriminates} between distilled tokens.

\paragraph{Probe C: prediction divergence under text ablation.}
For each cell we forecast twice from the same numeric input, once
with the row's text and once with the text branch ablated,
averaging ten samples from \Aurora{}'s flow-matching head per
setting:
\[
\widehat{D}
= \bbE_b \bigl\|\yhat_{\text{text}}(b) - \yhat_{\text{none}}(b)\bigr\|_2^2.
\]
This asks whether the text branch \emph{matters} for the forward
pass: $\widehat{D}\!\ll\!\MSE$ means the prediction is the same
with or without it.

\begin{table}[h]
\centering
\small
\setlength{\tabcolsep}{4pt}
\renewcommand{\arraystretch}{1.1}
\caption{\textbf{Full probe results}, mean $\pm$ s.d.\ over $n=27$
probe cells (9 domains $\times$ 3 seeds at fixed $H\!=\!8$). The
text-only conditions all show non-zero $\widehat{g}$ and uniformly
high $\widehat{H}$, with negligible $\widehat{D}$.}
\label{tab:probes_full}
\begin{tabular}{@{}lccc@{}}
\toprule
\textbf{Condition} & $\widehat{g}$ & $\widehat{H}$ & $\widehat{D}$ \\
\midrule
\Coriginal     & $0.152\pm0.240$ & $0.975\pm0.009$ & $0.041\pm0.037$ \\
\Cempty        & $0.074\pm0.103$ & $0.975\pm0.009$ & $0.040\pm0.041$ \\
\Cconst        & $0.156\pm0.228$ & $0.976\pm0.009$ & $0.044\pm0.035$ \\
\Cshuf         & $0.155\pm0.400$ & $0.975\pm0.009$ & $0.035\pm0.035$ \\
\Ccross        & $0.072\pm0.082$ & $0.975\pm0.009$ & $0.035\pm0.032$ \\
\Coracle       & $0.133\pm0.187$ & $0.976\pm0.009$ & $0.058\pm0.084$ \\
\midrule
\Czero{}       & $0.265\pm0.673$ & $0.975\pm0.009$ & $0.028\pm0.018$ \\
\Cflag{}       & $0.080\pm0.153$ & $0.975\pm0.009$ & $0.035\pm0.024$ \\
\bottomrule
\end{tabular}
\end{table}

\textit{Reading Table~\ref{tab:probes_full}.} The first column
($\widehat{g}$) is non-zero everywhere, ruling out a never-trained
pathway. The second column ($\widehat{H}$) sits at $0.975$ on every
row including the oracle, showing that the guider's attention is
nearly uniform across its distilled tokens regardless of text
content. The third column ($\widehat{D}$) is at most $0.058$,
which is two orders of magnitude below the test-MSE scale of
$\sim 8.6$, showing that disabling the text branch barely changes
the forecast. These three readings together describe a pathway
that is trained but content-blind in the forward pass.

% ================================================================== %
\section{Text-Side Diagnostics: Definitions and Full Results}
\label{sec:app:diagnostics}

We compute three diagnostics on each domain's per-row text
embeddings, separately for the GPT-2 and BERT encoders.

\paragraph{TTW (temporal text--target Wasserstein distance).}
Introduced by~\citet{li2025tats}, TTW measures alignment between
text-embedding trajectories and the target series. Let
$\mathbf{e}_t$ be the per-row text embedding and $y_t$ the target
at time $t$. Define the centred unit-norm embeddings
$\tilde{\mathbf{e}}_t = (\mathbf{e}_t - \bar{\mathbf{e}})/\|\mathbf{e}_t - \bar{\mathbf{e}}\|_2$
and the lag-similarity profile
\[
\mathrm{Sim}(k) = \frac{1}{T-k}\sum_{t=1}^{T-k}\,
                  \tilde{\mathbf{e}}_t^{\top}\tilde{\mathbf{e}}_{t+k}.
\]
Let $A_y$ be the $L_1$-normalised amplitude spectrum of
$\Delta y_t = y_{t+1} - y_t$, and $A_e$ the $L_1$-normalised
amplitude spectrum of $\Delta \mathrm{Sim}(k)$. Then
\[
\mathrm{TTW} = W_1(A_y,\, A_e),
\]
the $1$-D Wasserstein distance between the two spectra on a
common frequency grid. Lower TTW means the dominant frequency
content of the text-embedding trajectory matches the dominant
frequency content of the target. The TaTS paper argues low TTW
indicates text is suitable as a covariate for time-series fusion.

\paragraph{ETA (embedding temporal autocorrelation).}
Mean lag-$1$ autocorrelation across embedding dimensions:
\[
\mathrm{ETA} = \frac{1}{D}\sum_{d=1}^{D}
   \frac{\sum_t (e_{t,d}-\bar e_d)(e_{t+1,d}-\bar e_d)}
        {\sum_t (e_{t,d}-\bar e_d)^2}.
\]
High ETA means embeddings evolve smoothly over time; low ETA
means consecutive embeddings look like independent draws.

\paragraph{SDI (semantic diversity index).}
\[
\mathrm{SDI} = 1 - \frac{1}{T-1}\sum_{t=1}^{T-1}
   \cos\bigl(\mathbf{e}_t,\,\mathbf{e}_{t+1}\bigr).
\]
High SDI indicates that consecutive rows are semantically
distinct. Low SDI is a structural problem for any encoder that
reads the text: if every row's embedding is nearly identical to
its neighbours, no attention head can extract row-specific
content from the text alone.

\begin{table}[h]
\centering
\small
\setlength{\tabcolsep}{3pt}
\renewcommand{\arraystretch}{1.05}
\caption{\textbf{Per-domain diagnostics on the original text.}
Eight domains; Environment excluded as self-paired.
\textbf{Bold} marks the domain with the highest SDI per encoder.}
\label{tab:diag_full}
\begin{tabular}{@{}l ccc ccc@{}}
\toprule
& \multicolumn{3}{c}{\textbf{GPT-2}} & \multicolumn{3}{c}{\textbf{BERT}} \\
\cmidrule(lr){2-4}\cmidrule(lr){5-7}
\textbf{Domain} & TTW & ETA & SDI & TTW & ETA & SDI \\
\midrule
Agriculture & 0.028 & 0.436 & 0.011 & 0.028 & 0.399 & 0.045 \\
Climate     & 0.078 & 0.372 & 0.004 & 0.025 & 0.339 & 0.014 \\
Economy     & 0.030 & 0.426 & 0.009 & 0.014 & 0.331 & 0.038 \\
Energy      & 0.056 & 0.597 & 0.004 & 0.028 & 0.575 & 0.017 \\
Health      & 0.022 & 0.534 & 0.005 & 0.028 & 0.483 & 0.020 \\
Security    & 0.027 & 0.056 & 0.011 & 0.027 & 0.071 & 0.039 \\
SocialGood  & 0.155 & 0.904 & 0.006 & 0.092 & 0.777 & 0.021 \\
Traffic     & 0.054 & 0.058 & \textbf{0.013} & 0.058 & 0.107 & \textbf{0.050} \\
\midrule
\textbf{Mean} & \textbf{0.056} & \textbf{0.423} & \textbf{0.008}
              & \textbf{0.037} & \textbf{0.385} & \textbf{0.031} \\
\bottomrule
\end{tabular}
\end{table}

\textit{Reading Table~\ref{tab:diag_full}.} TTW is uniformly low
across all domains, indicating aggregate alignment exists. SDI is
also uniformly low: the highest GPT-2 SDI is $0.013$ on Traffic,
which means consecutive rows look nearly identical to the encoder.
ETA shows wide variation: SocialGood and Energy text evolves
smoothly (high ETA), while Security and Traffic look closer to
independent draws (low ETA). Even the most distinctive domains
do not reach the level of row-to-row contrast an attentive head
would need.

\paragraph{Diagnostics under text perturbations.}
Our perturbations are designed to manipulate these three
properties of the text. Table~\ref{tab:diag_perturb} measures
TTW, ETA, and SDI on each perturbed text column, with all five
substitutions producing dramatically different diagnostic
signatures. The takeaway from comparing
Table~\ref{tab:diag_perturb} to the main results is that despite
order-of-magnitude swings in these diagnostics, downstream MSE
moves by less than $0.5\%$ in every case.

\begin{table}[h]
\centering
\small
\setlength{\tabcolsep}{3.5pt}
\renewcommand{\arraystretch}{1.08}
\caption{\textbf{Diagnostics under text perturbations,} mean over
$8$ domains. Empty/Constant collapse SDI to $0$ (every row identical)
and push ETA toward $1$. Shuffled inflates SDI an order of
magnitude. Cross-domain text doubles TTW.
\textbf{Bold} marks values that differ from Original by a factor
$\geq 2$.}
\label{tab:diag_perturb}
\begin{tabular}{@{}l ccc ccc@{}}
\toprule
& \multicolumn{3}{c}{\textbf{GPT-2}} & \multicolumn{3}{c}{\textbf{BERT}} \\
\cmidrule(lr){2-4}\cmidrule(lr){5-7}
\textbf{Condition} & TTW & ETA & SDI & TTW & ETA & SDI \\
\midrule
Original     & 0.056 & 0.423 & 0.008 & 0.037 & 0.385 & 0.031 \\
Empty        & 0.054 & \textbf{0.820} & \textbf{0.000} & \textbf{0.158} & \textbf{0.978} & \textbf{0.000} \\
Constant     & 0.049 & \textbf{0.733} & \textbf{0.000} & \textbf{0.131} & \textbf{0.965} & \textbf{0.000} \\
Shuffled     & 0.060 & \textbf{0.000} & \textbf{0.044} & 0.063 & \textbf{0.000} & \textbf{0.084} \\
Cross-domain & \textbf{0.112} & 0.455 & 0.008 & \textbf{0.108} & 0.418 & 0.029 \\
\bottomrule
\end{tabular}
\end{table}

\textit{Reading Table~\ref{tab:diag_perturb}.} The four
text-content perturbations move the three diagnostics in
qualitatively different ways. Empty and Constant make every row
identical, so SDI collapses to zero and ETA approaches one.
Shuffled produces highly distinctive consecutive rows
(SDI on BERT increases from $0.031$ to $0.084$, an
order-of-magnitude rise) but with no temporal structure
(ETA falls to zero). Cross-domain text doubles TTW from $0.037$
to $0.108$ on BERT, breaking the alignment between text and
target trajectories. Despite all of these large diagnostic
swings, downstream MSE on every model and every backbone moves by
less than $0.5\%$ on the corresponding text-only conditions
(Tables~\ref{tab:full_aurora}, \ref{tab:full_tats}, \ref{tab:full_mmtsf}). On
\TimeMMD{}, low TTW does correspond to text being measurable
co-aligned with the target, in line with~\citet{li2025tats}, but
neither TTW nor ETA nor SDI individually predicts whether the
text is being \emph{used} by the trained model. We do not refute
the TTW criterion as a property of the text corpus; we observe
that on this benchmark, none of the three diagnostics is a
reliable proxy for whether a perturbation will move the model's
forecast.

% ================================================================== %
\section{Full Results: Per-Backbone, Per-Domain, Per-Condition}
\label{sec:app:fullresults}

This section gives the complete experiment grid for all three
architectures. For \TaTS{} and \MMTSF{} (eight backbones each), we
split the eight backbones into two tables of four each. \Aurora{}
has only one configuration. Each row group within a table fixes the
backbone; rows within a group vary the condition; columns are the
nine \TimeMMD{} domains plus an aggregate average. The
\textbf{Orig.} row gives absolute mean test MSE for the
original-text baseline; subsequent rows give the percent change
relative to that baseline within each domain.

We use a single visual convention throughout. \textbf{Bold} marks
deltas with $|\Delta|\!\geq\!0.5\%$ (a substantive movement at this
benchmark's noise floor). Grey font marks $|\Delta|\!<\!0.05\%$
(numerically indistinguishable from zero at our reporting
precision). All other deltas are rendered in normal weight.

The dominant pattern is visible at a glance: the rows for the five
text-content perturbations (Empty, Const., Shuf., Cross, Oracle)
sit in grey or near-grey across all backbones and all domains.
The two structural-perturbation rows ($\pnum{=}0$ and Unimod.)
are filled with bold deltas. Per-domain noise on small-baseline
domains (e.g.\ Economy on \Aurora{}, where Original MSE is
$0.033$) can produce isolated bold cells in text-only rows; these
sit at one or two per backbone and reflect floating-point and
sampling noise on a tiny base, not a real text effect.

\begin{table*}[t]
\centering
\scriptsize
\setlength{\tabcolsep}{2.5pt}
\renewcommand{\arraystretch}{1.0}
\providecommand{\sm}[1]{\textcolor{gray!50}{#1}}
\caption{\textbf{Full \Aurora{} results.} Mean MSE on the original-text baseline (\textbf{Orig.} row); subsequent rows give $\Delta\%$ MSE relative to that baseline. $\pnum{=}0$ zeroes the numeric column $\pnum$ with text intact. \textbf{Unimod.} disables the text branch (\Aurora{}'s unimodal baseline). \Aurora{} does not consume $\pnum$, so the $\pnum{=}0$ row matches Original. Bold marks $|\Delta|\geq 0.5\%$; grey marks $|\Delta|<0.05\%$.}
\label{tab:full_aurora}
\begin{tabular}{@{}llcccccccccc@{}}
\toprule
\textbf{Backbone} & \textbf{Cond.} & \textbf{Agri} & \textbf{Clim} & \textbf{Econ} & \textbf{Ener} & \textbf{Envi} & \textbf{Heal} & \textbf{Secu} & \textbf{SocG} & \textbf{Traf} & \textbf{Avg.} \\
\midrule
\multirow{8}{*}{\textsc{Aurora}} & Orig. & 0.275 & 0.865 & 0.033 & 0.255 & 0.276 & 1.55 & 72.7 & 0.836 & 0.161 & \textbf{8.55} \\
 & Empty & $+0.09$ & \sm{$+0.01$} & $\mathbf{-3.47}$ & $+0.36$ & \sm{$-0.04$} & $-0.10$ & \sm{$+0.03$} & \sm{$-0.03$} & $-0.20$ & $+0.02$ \\
 & Const. & $\mathbf{+0.64}$ & \sm{$+0.03$} & $\mathbf{+4.70}$ & \sm{$+0.05$} & $+0.06$ & $+0.39$ & \sm{$0.00$} & $\mathbf{+1.05}$ & $\mathbf{+1.05}$ & $+0.03$ \\
 & Shuf. & \sm{$+0.01$} & \sm{$+0.01$} & $\mathbf{+2.21}$ & \sm{$+0.01$} & \sm{$0.00$} & \sm{$0.00$} & \sm{$0.00$} & $-0.08$ & $+0.20$ & \sm{$0.00$} \\
 & Cross & $+0.09$ & \sm{$+0.02$} & $\mathbf{+0.78}$ & \sm{$+0.02$} & \sm{$0.00$} & \sm{$-0.05$} & \sm{$-0.01$} & $+0.32$ & $-0.25$ & $-0.01$ \\
 & Oracle & \sm{$+0.04$} & \sm{$+0.02$} & $\mathbf{+4.13}$ & $+0.25$ & \sm{$0.00$} & $+0.27$ & \sm{$0.00$} & $\mathbf{+0.93}$ & $\mathbf{+0.88}$ & $+0.02$ \\
 & $\pnum{=}0$ & \sm{$0.00$} & \sm{$0.00$} & \sm{$0.00$} & \sm{$0.00$} & \sm{$0.00$} & \sm{$0.00$} & \sm{$0.00$} & \sm{$0.00$} & \sm{$0.00$} & \sm{$0.00$} \\
 & Unimod. & \sm{$0.00$} & \sm{$0.00$} & \sm{$0.00$} & \sm{$0.00$} & \sm{$0.00$} & \sm{$0.00$} & \sm{$0.00$} & \sm{$0.00$} & \sm{$0.00$} & \sm{$0.00$} \\
\bottomrule
\end{tabular}
\end{table*}

\begin{table*}[t]
\centering
\scriptsize
\setlength{\tabcolsep}{2.5pt}
\renewcommand{\arraystretch}{0.96}
\providecommand{\sm}[1]{\textcolor{gray!50}{#1}}
\caption{\textbf{Full \TaTS{} results} across eight backbones.
\textbf{Orig.} row gives the mean test MSE on the original-text
baseline; all rows below it report the percentage change in MSE
relative to the \textbf{Orig.} row of that same backbone, within each
domain. \textbf{Bold} marks $|\Delta|\!\geq\!0.5\%$; grey marks
$|\Delta|\!<\!0.05\%$. Note: $\pnum{=}0$ on \TaTS{} carries the
amplitude artifact described in the caption of Table~\ref{tab:prior}.}
\label{tab:full_tats}
\begin{tabular}{@{}llcccccccccc@{}}
\toprule
\textbf{Backbone} & \textbf{Cond.} & \textbf{Agri} & \textbf{Clim} & \textbf{Econ} & \textbf{Ener} & \textbf{Envi} & \textbf{Heal} & \textbf{Secu} & \textbf{SocG} & \textbf{Traf} & \textbf{Avg.} \\
\midrule
\multirow{8}{*}{\textsc{Autoformer}} & Orig. & 0.131 & 0.987 & 0.039 & 0.407 & 0.315 & 1.53 & 107.3 & 1.15 & 0.187 & \textbf{12.45} \\
 & Empty & \sm{$0.00$} & \sm{$+0.02$} & \sm{$0.00$} & $\mathbf{+1.03}$ & $-0.26$ & $+0.08$ & \sm{$0.00$} & \sm{$0.00$} & \sm{$0.00$} & \sm{$0.00$} \\
 & Const. & \sm{$0.00$} & \sm{$0.00$} & \sm{$0.00$} & \sm{$0.00$} & $\mathbf{-0.82}$ & \sm{$0.00$} & \sm{$0.00$} & \sm{$0.00$} & \sm{$0.00$} & \sm{$0.00$} \\
 & Shuf. & \sm{$0.00$} & \sm{$0.00$} & \sm{$0.00$} & \sm{$+0.01$} & $-0.07$ & \sm{$0.00$} & \sm{$0.00$} & \sm{$0.00$} & \sm{$0.00$} & \sm{$0.00$} \\
 & Cross & \sm{$0.00$} & \sm{$+0.02$} & \sm{$0.00$} & $\mathbf{+1.03}$ & $\mathbf{-1.24}$ & \sm{$0.00$} & \sm{$0.00$} & \sm{$0.00$} & \sm{$0.00$} & \sm{$0.00$} \\
 & Oracle & \sm{$0.00$} & \sm{$0.00$} & \sm{$0.00$} & $\mathbf{+1.13}$ & $-0.20$ & $-0.09$ & \sm{$0.00$} & \sm{$0.00$} & \sm{$0.00$} & \sm{$0.00$} \\
 & $\pnum{=}0$ & $\mathbf{+553}$ & $\mathbf{+22.0}$ & $\mathbf{+633}$ & $\mathbf{+17.5}$ & $\mathbf{+83.8}$ & $\mathbf{+22.6}$ & $\mathbf{+2.75}$ & $\mathbf{-3.41}$ & $\mathbf{+131}$ & $\mathbf{+4.48}$ \\
 & Unimod. & $\mathbf{-13.9}$ & $\mathbf{+31.5}$ & $\mathbf{+132}$ & $\mathbf{-11.1}$ & $\mathbf{+33.5}$ & $\mathbf{+33.9}$ & $\mathbf{+14.8}$ & $\mathbf{-7.14}$ & $\mathbf{+23.3}$ & $\mathbf{+15.0}$ \\
\midrule
\multirow{8}{*}{\textsc{Crossformer}} & Orig. & 0.202 & 1.00 & 0.231 & 0.383 & 0.297 & 1.18 & 122.4 & 0.924 & 0.177 & \textbf{14.09} \\
 & Empty & \sm{$0.00$} & \sm{$0.00$} & \sm{$0.00$} & \sm{$0.00$} & \sm{$0.00$} & \sm{$0.00$} & \sm{$0.00$} & \sm{$0.00$} & \sm{$0.00$} & \sm{$0.00$} \\
 & Const. & \sm{$0.00$} & \sm{$0.00$} & \sm{$0.00$} & \sm{$0.00$} & \sm{$0.00$} & \sm{$0.00$} & \sm{$0.00$} & \sm{$0.00$} & \sm{$0.00$} & \sm{$0.00$} \\
 & Shuf. & \sm{$0.00$} & \sm{$0.00$} & \sm{$0.00$} & \sm{$0.00$} & \sm{$0.00$} & \sm{$0.00$} & \sm{$0.00$} & \sm{$0.00$} & \sm{$0.00$} & \sm{$0.00$} \\
 & Cross & \sm{$0.00$} & \sm{$0.00$} & \sm{$0.00$} & \sm{$0.00$} & \sm{$0.00$} & \sm{$0.00$} & \sm{$0.00$} & \sm{$0.00$} & \sm{$0.00$} & \sm{$0.00$} \\
 & Oracle & \sm{$0.00$} & \sm{$0.00$} & \sm{$0.00$} & \sm{$0.00$} & \sm{$0.00$} & \sm{$0.00$} & \sm{$0.00$} & \sm{$0.00$} & \sm{$0.00$} & \sm{$0.00$} \\
 & $\pnum{=}0$ & $\mathbf{+199}$ & $\mathbf{+12.2}$ & $\mathbf{+484}$ & $\mathbf{-19.8}$ & $\mathbf{+52.3}$ & $\mathbf{+20.7}$ & $\mathbf{+3.71}$ & $\mathbf{-5.97}$ & $\mathbf{+45.2}$ & $\mathbf{+5.15}$ \\
 & Unimod. & $\mathbf{+109}$ & $\mathbf{+16.7}$ & $\mathbf{+293}$ & $\mathbf{-19.2}$ & $\mathbf{+53.9}$ & $\mathbf{+17.8}$ & $\mathbf{+3.10}$ & $\mathbf{-4.74}$ & $\mathbf{+41.4}$ & $\mathbf{+4.09}$ \\
\midrule
\multirow{8}{*}{\textsc{DLinear}} & Orig. & 0.154 & 0.959 & 0.042 & 0.466 & 0.322 & 1.55 & 106.9 & 1.15 & 0.209 & \textbf{12.42} \\
 & Empty & \sm{$0.00$} & \sm{$0.00$} & \sm{$0.00$} & \sm{$0.00$} & \sm{$0.00$} & \sm{$0.00$} & \sm{$0.00$} & \sm{$0.00$} & \sm{$0.00$} & \sm{$0.00$} \\
 & Const. & \sm{$0.00$} & \sm{$0.00$} & \sm{$0.00$} & \sm{$0.00$} & \sm{$0.00$} & \sm{$0.00$} & \sm{$0.00$} & \sm{$0.00$} & \sm{$0.00$} & \sm{$0.00$} \\
 & Shuf. & \sm{$0.00$} & \sm{$0.00$} & \sm{$0.00$} & \sm{$0.00$} & \sm{$0.00$} & \sm{$0.00$} & \sm{$0.00$} & \sm{$0.00$} & \sm{$0.00$} & \sm{$0.00$} \\
 & Cross & \sm{$0.00$} & \sm{$0.00$} & \sm{$0.00$} & \sm{$0.00$} & \sm{$0.00$} & \sm{$0.00$} & \sm{$0.00$} & \sm{$0.00$} & \sm{$0.00$} & \sm{$0.00$} \\
 & Oracle & \sm{$0.00$} & \sm{$0.00$} & \sm{$0.00$} & \sm{$0.00$} & \sm{$0.00$} & \sm{$0.00$} & \sm{$0.00$} & \sm{$0.00$} & \sm{$0.00$} & \sm{$0.00$} \\
 & $\pnum{=}0$ & $\mathbf{+16273}$ & $\mathbf{+663}$ & $\mathbf{+192}$ & $\mathbf{+514}$ & $\mathbf{+54.7}$ & $\mathbf{+82.8}$ & $\mathbf{+7.65}$ & $\mathbf{+265}$ & $\mathbf{+4243}$ & $\mathbf{+49.6}$ \\
 & Unimod. & $\mathbf{+39.6}$ & $\mathbf{+45.3}$ & $\mathbf{+236}$ & $\mathbf{-17.8}$ & $\mathbf{+72.5}$ & $\mathbf{+27.7}$ & $\mathbf{+4.97}$ & $\mathbf{+1.80}$ & $\mathbf{+90.5}$ & $\mathbf{+5.99}$ \\
\midrule
\multirow{8}{*}{\textsc{FEDformer}} & Orig. & 0.112 & 0.945 & 0.021 & 0.443 & 0.283 & 1.38 & 107.9 & 1.07 & 0.170 & \textbf{12.48} \\
 & Empty & \sm{$0.00$} & \sm{$0.00$} & \sm{$0.00$} & \sm{$0.00$} & \sm{$0.00$} & \sm{$0.00$} & \sm{$0.00$} & \sm{$0.00$} & \sm{$0.00$} & \sm{$0.00$} \\
 & Const. & \sm{$0.00$} & \sm{$0.00$} & \sm{$0.00$} & \sm{$0.00$} & \sm{$0.00$} & \sm{$0.00$} & \sm{$0.00$} & \sm{$0.00$} & \sm{$0.00$} & \sm{$0.00$} \\
 & Shuf. & \sm{$0.00$} & \sm{$0.00$} & \sm{$0.00$} & \sm{$0.00$} & \sm{$0.00$} & \sm{$0.00$} & \sm{$0.00$} & \sm{$0.00$} & \sm{$0.00$} & \sm{$0.00$} \\
 & Cross & \sm{$0.00$} & \sm{$0.00$} & \sm{$0.00$} & \sm{$0.00$} & $-0.10$ & \sm{$0.00$} & \sm{$0.00$} & \sm{$0.00$} & \sm{$0.00$} & \sm{$0.00$} \\
 & Oracle & \sm{$0.00$} & \sm{$0.00$} & \sm{$0.00$} & \sm{$0.00$} & \sm{$0.00$} & \sm{$0.00$} & \sm{$0.00$} & \sm{$0.00$} & \sm{$0.00$} & \sm{$0.00$} \\
 & $\pnum{=}0$ & $\mathbf{+306}$ & $\mathbf{+19.0}$ & $\mathbf{+6287}$ & $\mathbf{+26.1}$ & $\mathbf{+31.6}$ & $\mathbf{+9.00}$ & $\mathbf{+2.52}$ & $\mathbf{+4.87}$ & $\mathbf{+123}$ & $\mathbf{+4.57}$ \\
 & Unimod. & $\mathbf{-12.9}$ & $\mathbf{+35.1}$ & $\mathbf{+183}$ & $\mathbf{-33.2}$ & $\mathbf{+38.0}$ & $\mathbf{+5.66}$ & $\mathbf{+7.62}$ & $\mathbf{-13.6}$ & $\mathbf{+26.8}$ & $\mathbf{+7.58}$ \\
\midrule
\multirow{8}{*}{\textsc{FiLM}} & Orig. & 0.120 & 0.956 & 0.015 & 0.413 & 0.270 & 1.53 & 108.7 & 1.09 & 0.179 & \textbf{12.59} \\
 & Empty & \sm{$0.00$} & \sm{$0.00$} & \sm{$0.00$} & \sm{$0.00$} & \sm{$0.00$} & \sm{$0.00$} & \sm{$0.00$} & \sm{$0.00$} & \sm{$0.00$} & \sm{$0.00$} \\
 & Const. & \sm{$0.00$} & \sm{$0.00$} & \sm{$0.00$} & \sm{$0.00$} & \sm{$0.00$} & \sm{$0.00$} & \sm{$0.00$} & \sm{$0.00$} & \sm{$0.00$} & \sm{$0.00$} \\
 & Shuf. & \sm{$0.00$} & \sm{$0.00$} & \sm{$0.00$} & \sm{$0.00$} & \sm{$0.00$} & \sm{$0.00$} & \sm{$0.00$} & \sm{$0.00$} & \sm{$0.00$} & \sm{$0.00$} \\
 & Cross & \sm{$0.00$} & \sm{$0.00$} & \sm{$0.00$} & \sm{$0.00$} & \sm{$0.00$} & \sm{$0.00$} & \sm{$0.00$} & \sm{$0.00$} & \sm{$0.00$} & \sm{$0.00$} \\
 & Oracle & \sm{$0.00$} & \sm{$0.00$} & \sm{$0.00$} & \sm{$0.00$} & \sm{$0.00$} & \sm{$0.00$} & \sm{$0.00$} & \sm{$0.00$} & \sm{$0.00$} & \sm{$0.00$} \\
 & $\pnum{=}0$ & $\mathbf{+21010}$ & $\mathbf{+670}$ & $\mathbf{+465}$ & $\mathbf{+703}$ & $\mathbf{+99.8}$ & $\mathbf{+78.3}$ & $\mathbf{+7.53}$ & $\mathbf{+294}$ & $\mathbf{+5069}$ & $\mathbf{+49.9}$ \\
 & Unimod. & $\mathbf{-11.1}$ & $\mathbf{+39.9}$ & $\mathbf{+125}$ & $\mathbf{-10.4}$ & $\mathbf{+19.7}$ & $\mathbf{+30.9}$ & $\mathbf{+9.42}$ & $\mathbf{-0.88}$ & $\mathbf{+49.6}$ & $\mathbf{+9.88}$ \\
\midrule
\multirow{8}{*}{\textsc{Informer}} & Orig. & 0.239 & 0.948 & 0.328 & 0.393 & 0.287 & 1.15 & 123.3 & 0.875 & 0.162 & \textbf{14.19} \\
 & Empty & \sm{$-0.01$} & \sm{$-0.02$} & \sm{$+0.03$} & $+0.15$ & $-0.12$ & $-0.31$ & \sm{$-0.01$} & \sm{$-0.02$} & \sm{$+0.01$} & $-0.01$ \\
 & Const. & $+0.05$ & \sm{$+0.04$} & \sm{$+0.01$} & $+0.12$ & $+0.16$ & \sm{$+0.03$} & \sm{$0.00$} & \sm{$+0.02$} & \sm{$+0.02$} & \sm{$0.00$} \\
 & Shuf. & $+0.10$ & \sm{$0.00$} & \sm{$0.00$} & $+0.12$ & $+0.05$ & $+0.09$ & \sm{$0.00$} & \sm{$+0.01$} & \sm{$+0.03$} & \sm{$0.00$} \\
 & Cross & \sm{$+0.04$} & \sm{$-0.01$} & \sm{$-0.03$} & $+0.08$ & $+0.26$ & $+0.09$ & \sm{$0.00$} & \sm{$+0.03$} & \sm{$-0.02$} & \sm{$0.00$} \\
 & Oracle & $+0.13$ & \sm{$-0.02$} & \sm{$+0.02$} & $+0.17$ & $\mathbf{+0.58}$ & $-0.16$ & \sm{$0.00$} & \sm{$-0.02$} & \sm{$+0.01$} & \sm{$0.00$} \\
 & $\pnum{=}0$ & $\mathbf{+321}$ & $\mathbf{+9.64}$ & $\mathbf{+243}$ & $\mathbf{-1.83}$ & $\mathbf{+52.2}$ & $\mathbf{+8.61}$ & $\mathbf{+6.21}$ & $+0.41$ & $\mathbf{+44.7}$ & $\mathbf{+7.54}$ \\
 & Unimod. & $\mathbf{+124}$ & $\mathbf{+20.3}$ & $\mathbf{+327}$ & $\mathbf{-1.77}$ & $\mathbf{+56.2}$ & $\mathbf{+23.9}$ & $\mathbf{+6.98}$ & $\mathbf{+1.18}$ & $\mathbf{+53.1}$ & $\mathbf{+8.38}$ \\
\midrule
\multirow{8}{*}{\textsc{Transformer}} & Orig. & 0.180 & 0.935 & 0.131 & 0.361 & 0.276 & 1.14 & 121.7 & 0.894 & 0.163 & \textbf{13.97} \\
 & Empty & \sm{$0.00$} & \sm{$0.00$} & \sm{$0.00$} & \sm{$0.00$} & \sm{$0.00$} & \sm{$0.00$} & \sm{$0.00$} & \sm{$0.00$} & \sm{$0.00$} & \sm{$0.00$} \\
 & Const. & \sm{$0.00$} & \sm{$0.00$} & \sm{$0.00$} & \sm{$0.00$} & \sm{$0.00$} & \sm{$0.00$} & \sm{$0.00$} & \sm{$0.00$} & \sm{$0.00$} & \sm{$0.00$} \\
 & Shuf. & \sm{$0.00$} & \sm{$0.00$} & \sm{$0.00$} & \sm{$0.00$} & \sm{$0.00$} & \sm{$0.00$} & \sm{$0.00$} & \sm{$0.00$} & \sm{$0.00$} & \sm{$0.00$} \\
 & Cross & \sm{$0.00$} & \sm{$0.00$} & \sm{$0.00$} & \sm{$0.00$} & \sm{$0.00$} & \sm{$0.00$} & \sm{$0.00$} & \sm{$0.00$} & \sm{$0.00$} & \sm{$0.00$} \\
 & Oracle & \sm{$0.00$} & \sm{$0.00$} & \sm{$0.00$} & \sm{$0.00$} & \sm{$0.00$} & \sm{$0.00$} & \sm{$0.00$} & \sm{$0.00$} & \sm{$0.00$} & \sm{$0.00$} \\
 & $\pnum{=}0$ & $\mathbf{+251}$ & $\mathbf{+14.9}$ & $\mathbf{+470}$ & $\mathbf{-20.3}$ & $\mathbf{+42.7}$ & $\mathbf{+7.12}$ & $\mathbf{+4.71}$ & $\mathbf{+3.05}$ & $\mathbf{+59.6}$ & $\mathbf{+5.71}$ \\
 & Unimod. & $\mathbf{+85.0}$ & $\mathbf{+18.4}$ & $\mathbf{+501}$ & $\mathbf{-14.4}$ & $\mathbf{+39.6}$ & $\mathbf{+16.9}$ & $\mathbf{+8.14}$ & $\mathbf{-1.22}$ & $\mathbf{+51.0}$ & $\mathbf{+8.91}$ \\
\midrule
\multirow{8}{*}{\textsc{iTransformer}} & Orig. & 0.094 & 0.996 & 0.010 & 0.302 & 0.260 & 1.30 & 115.5 & 1.07 & 0.195 & \textbf{13.31} \\
 & Empty & \sm{$0.00$} & \sm{$0.00$} & \sm{$0.00$} & \sm{$0.00$} & \sm{$0.00$} & \sm{$0.00$} & \sm{$0.00$} & \sm{$0.00$} & \sm{$0.00$} & \sm{$0.00$} \\
 & Const. & \sm{$0.00$} & \sm{$0.00$} & \sm{$0.00$} & \sm{$0.00$} & \sm{$0.00$} & \sm{$0.00$} & \sm{$0.00$} & \sm{$0.00$} & \sm{$0.00$} & \sm{$0.00$} \\
 & Shuf. & \sm{$0.00$} & \sm{$0.00$} & \sm{$0.00$} & \sm{$0.00$} & \sm{$0.00$} & \sm{$0.00$} & \sm{$0.00$} & \sm{$0.00$} & \sm{$0.00$} & \sm{$0.00$} \\
 & Cross & \sm{$0.00$} & \sm{$0.00$} & \sm{$0.00$} & \sm{$0.00$} & \sm{$0.00$} & \sm{$0.00$} & \sm{$0.00$} & \sm{$0.00$} & \sm{$0.00$} & \sm{$0.00$} \\
 & Oracle & \sm{$0.00$} & \sm{$0.00$} & \sm{$0.00$} & \sm{$0.00$} & \sm{$0.00$} & \sm{$0.00$} & \sm{$0.00$} & \sm{$0.00$} & \sm{$0.00$} & \sm{$0.00$} \\
 & $\pnum{=}0$ & $\mathbf{+23169}$ & $\mathbf{+60.9}$ & $\mathbf{+210}$ & $\mathbf{+251}$ & $\mathbf{+10.1}$ & $\mathbf{+52.3}$ & $\mathbf{+13.3}$ & $\mathbf{+348}$ & $\mathbf{+1597}$ & $\mathbf{+38.4}$ \\
 & Unimod. & $\mathbf{-1.65}$ & $\mathbf{+21.9}$ & $\mathbf{+60.5}$ & $\mathbf{-11.6}$ & $\mathbf{+6.91}$ & $\mathbf{+28.5}$ & $\mathbf{+0.63}$ & $\mathbf{+9.75}$ & $\mathbf{+13.0}$ & $\mathbf{+1.20}$ \\
\bottomrule
\end{tabular}
\end{table*}

\begin{table*}[t]
\centering
\scriptsize
\setlength{\tabcolsep}{2.5pt}
\renewcommand{\arraystretch}{0.96}
\providecommand{\sm}[1]{\textcolor{gray!50}{#1}}
\caption{\textbf{Full \MMTSF{} results} across eight backbones.
\textbf{Orig.} row gives the mean test MSE on the original-text
baseline; all rows below it report the percentage change in MSE
relative to the \textbf{Orig.} row of that same backbone, within each
domain. \textbf{Bold} marks $|\Delta|\!\geq\!0.5\%$; grey marks
$|\Delta|\!<\!0.05\%$.}
\label{tab:full_mmtsf}
\begin{tabular}{@{}llcccccccccc@{}}
\toprule
\textbf{Backbone} & \textbf{Cond.} & \textbf{Agri} & \textbf{Clim} & \textbf{Econ} & \textbf{Ener} & \textbf{Envi} & \textbf{Heal} & \textbf{Secu} & \textbf{SocG} & \textbf{Traf} & \textbf{Avg.} \\
\midrule
\multirow{8}{*}{\textsc{Autoformer}} & Orig. & 0.105 & 1.19 & 0.072 & 0.364 & 0.567 & 1.90 & 122.1 & 1.06 & 0.213 & \textbf{14.17} \\
 & Empty & $\mathbf{+3.49}$ & $+0.06$ & $\mathbf{+0.63}$ & $\mathbf{-3.20}$ & $-0.08$ & $-0.39$ & $-0.29$ & $\mathbf{-1.12}$ & $\mathbf{-2.60}$ & $-0.31$ \\
 & Const. & $\mathbf{+1.79}$ & $-0.30$ & $\mathbf{+9.07}$ & \sm{$-0.01$} & $\mathbf{-1.31}$ & $\mathbf{-1.29}$ & $\mathbf{+0.62}$ & $\mathbf{-1.69}$ & $\mathbf{-0.80}$ & $\mathbf{+0.56}$ \\
 & Shuf. & $\mathbf{-2.08}$ & $\mathbf{+0.97}$ & $\mathbf{-0.80}$ & $+0.46$ & $-0.19$ & $\mathbf{+1.68}$ & $-0.26$ & $\mathbf{+2.33}$ & $\mathbf{-1.85}$ & $-0.20$ \\
 & Cross & $\mathbf{-2.90}$ & $\mathbf{+1.46}$ & $\mathbf{-12.8}$ & $\mathbf{+2.76}$ & $-0.46$ & $\mathbf{+0.88}$ & $-0.22$ & $\mathbf{+1.22}$ & $-0.40$ & $-0.18$ \\
 & Oracle & $\mathbf{+5.76}$ & \sm{$-0.02$} & $-0.41$ & $-0.41$ & $-0.12$ & $\mathbf{+0.85}$ & $\mathbf{-2.45}$ & $\mathbf{-0.88}$ & $\mathbf{-2.46}$ & $\mathbf{-2.34}$ \\
 & $\pnum{=}0$ & $\mathbf{+46.8}$ & $\mathbf{+7.64}$ & $\mathbf{+70.2}$ & $\mathbf{-4.26}$ & $\mathbf{+3.25}$ & $\mathbf{+5.50}$ & $-0.17$ & $\mathbf{-2.73}$ & $\mathbf{-0.57}$ & $+0.05$ \\
 & Unimod. & $+0.49$ & $\mathbf{+8.94}$ & $\mathbf{+5.71}$ & $\mathbf{-1.33}$ & $\mathbf{+5.84}$ & $\mathbf{+8.66}$ & $\mathbf{+3.68}$ & $\mathbf{-1.01}$ & $\mathbf{+3.99}$ & $\mathbf{+3.76}$ \\
\midrule
\multirow{8}{*}{\textsc{Crossformer}} & Orig. & 0.309 & 1.12 & 0.729 & 0.331 & 0.544 & 1.31 & 126.9 & 0.884 & 0.218 & \textbf{14.70} \\
 & Empty & $\mathbf{-2.92}$ & $\mathbf{+0.51}$ & $\mathbf{-4.03}$ & $+0.41$ & $\mathbf{+2.28}$ & $+0.13$ & \sm{$+0.02$} & $\mathbf{-0.92}$ & $\mathbf{-0.61}$ & \sm{$0.00$} \\
 & Const. & $\mathbf{-2.95}$ & $\mathbf{+0.51}$ & $\mathbf{-4.08}$ & $+0.41$ & $\mathbf{+0.58}$ & $+0.13$ & \sm{$+0.02$} & $\mathbf{-0.92}$ & $\mathbf{-0.62}$ & $-0.01$ \\
 & Shuf. & $\mathbf{+1.30}$ & $-0.14$ & $\mathbf{-0.98}$ & \sm{$0.00$} & $\mathbf{+1.09}$ & \sm{$0.00$} & \sm{$0.00$} & \sm{$-0.02$} & \sm{$0.00$} & \sm{$0.00$} \\
 & Cross & $\mathbf{+1.32}$ & $\mathbf{-0.97}$ & $\mathbf{-0.94}$ & $\mathbf{+0.79}$ & $+0.43$ & $+0.26$ & $+0.06$ & \sm{$+0.02$} & $-0.26$ & $+0.06$ \\
 & Oracle & $\mathbf{+0.95}$ & \sm{$+0.04$} & $\mathbf{-3.94}$ & $+0.29$ & $\mathbf{+0.80}$ & $\mathbf{+1.65}$ & $+0.21$ & $\mathbf{-1.04}$ & $\mathbf{-2.50}$ & $+0.19$ \\
 & $\pnum{=}0$ & $\mathbf{+25.4}$ & $\mathbf{+4.31}$ & $\mathbf{+31.5}$ & $\mathbf{+2.62}$ & $\mathbf{+8.66}$ & $\mathbf{+4.78}$ & $\mathbf{+0.70}$ & $\mathbf{+2.24}$ & $\mathbf{+6.98}$ & $\mathbf{+1.06}$ \\
 & Unimod. & $\mathbf{+12.1}$ & $\mathbf{+4.04}$ & $\mathbf{+20.4}$ & $\mathbf{+1.04}$ & $\mathbf{+8.08}$ & $\mathbf{+4.04}$ & $\mathbf{+0.77}$ & $\mathbf{+1.83}$ & $\mathbf{+8.26}$ & $\mathbf{+1.02}$ \\
\midrule
\multirow{8}{*}{\textsc{DLinear}} & Orig. & 0.155 & 1.23 & 0.072 & 0.362 & 0.536 & 1.70 & 109.3 & 1.09 & 0.303 & \textbf{12.75} \\
 & Empty & \sm{$-0.04$} & \sm{$0.00$} & \sm{$+0.05$} & \sm{$0.00$} & $\mathbf{+0.96}$ & \sm{$0.00$} & \sm{$0.00$} & \sm{$+0.01$} & \sm{$-0.01$} & $+0.01$ \\
 & Const. & $-0.05$ & \sm{$0.00$} & $\mathbf{-0.51}$ & \sm{$0.00$} & $\mathbf{+0.96}$ & \sm{$0.00$} & \sm{$0.00$} & \sm{$+0.01$} & \sm{$-0.03$} & \sm{$0.00$} \\
 & Shuf. & \sm{$0.00$} & \sm{$+0.01$} & $-0.48$ & $+0.21$ & $\mathbf{+0.95}$ & \sm{$0.00$} & \sm{$0.00$} & \sm{$+0.01$} & \sm{$-0.02$} & $+0.01$ \\
 & Cross & \sm{$-0.04$} & \sm{$0.00$} & \sm{$+0.04$} & $+0.22$ & $\mathbf{+0.96}$ & \sm{$+0.01$} & \sm{$0.00$} & \sm{$0.00$} & \sm{$-0.01$} & $+0.01$ \\
 & Oracle & \sm{$-0.03$} & \sm{$+0.01$} & \sm{$+0.02$} & $+0.22$ & $\mathbf{+0.93}$ & $+0.15$ & \sm{$0.00$} & \sm{$+0.01$} & $-0.05$ & $+0.01$ \\
 & $\pnum{=}0$ & $\mathbf{+745}$ & $\mathbf{+20.1}$ & $\mathbf{+6.94}$ & $\mathbf{+31.0}$ & $\mathbf{-5.54}$ & $\mathbf{+9.58}$ & $\mathbf{+0.65}$ & $\mathbf{-1.85}$ & $\mathbf{+110}$ & $\mathbf{+2.33}$ \\
 & Unimod. & $\mathbf{+38.1}$ & $\mathbf{+13.1}$ & $\mathbf{+95.0}$ & $\mathbf{+0.96}$ & $\mathbf{+9.07}$ & $\mathbf{+4.08}$ & $\mathbf{+2.20}$ & $\mathbf{+6.94}$ & $\mathbf{+30.0}$ & $\mathbf{+2.60}$ \\
\midrule
\multirow{8}{*}{\textsc{FEDformer}} & Orig. & 0.093 & 1.15 & 0.045 & 0.267 & 0.493 & 1.41 & 115.7 & 0.946 & 0.180 & \textbf{13.36} \\
 & Empty & $\mathbf{+2.23}$ & $\mathbf{-0.83}$ & $\mathbf{+10.8}$ & $\mathbf{+2.20}$ & $+0.26$ & $-0.43$ & $\mathbf{+0.64}$ & \sm{$+0.02$} & $+0.19$ & $\mathbf{+0.61}$ \\
 & Const. & $\mathbf{+2.21}$ & $\mathbf{-0.83}$ & $\mathbf{+10.9}$ & $\mathbf{+2.19}$ & $+0.25$ & $-0.44$ & $\mathbf{+0.65}$ & \sm{$+0.03$} & $+0.19$ & $\mathbf{+0.62}$ \\
 & Shuf. & $\mathbf{-1.31}$ & \sm{$+0.03$} & $\mathbf{+8.09}$ & \sm{$0.00$} & $+0.14$ & \sm{$+0.01$} & \sm{$-0.01$} & $\mathbf{+1.11}$ & \sm{$+0.01$} & \sm{$0.00$} \\
 & Cross & $\mathbf{-3.35}$ & $-0.28$ & $+0.15$ & $\mathbf{+2.92}$ & $-0.06$ & $+0.11$ & $-0.12$ & $\mathbf{+0.53}$ & $-0.32$ & $-0.11$ \\
 & Oracle & $\mathbf{+1.27}$ & $-0.07$ & $\mathbf{-1.41}$ & $-0.50$ & \sm{$+0.02$} & $-0.36$ & $\mathbf{+0.86}$ & $\mathbf{-0.59}$ & $\mathbf{-0.65}$ & $\mathbf{+0.82}$ \\
 & $\pnum{=}0$ & $\mathbf{+39.4}$ & $\mathbf{+0.85}$ & $\mathbf{+136}$ & $\mathbf{+5.87}$ & $\mathbf{+3.73}$ & $+0.43$ & $-0.15$ & $\mathbf{-1.36}$ & $\mathbf{+3.97}$ & $-0.03$ \\
 & Unimod. & $\mathbf{+3.16}$ & $\mathbf{+4.65}$ & $\mathbf{+34.1}$ & $\mathbf{+1.83}$ & $\mathbf{+7.11}$ & $\mathbf{+3.23}$ & $+0.13$ & \sm{$0.00$} & $\mathbf{+4.29}$ & $+0.27$ \\
\midrule
\multirow{8}{*}{\textsc{FiLM}} & Orig. & 0.106 & 1.24 & 0.034 & 0.362 & 0.506 & 1.80 & 116.6 & 1.05 & 0.238 & \textbf{13.55} \\
 & Empty & $+0.32$ & \sm{$-0.03$} & $-0.05$ & $+0.44$ & $-0.05$ & \sm{$+0.01$} & \sm{$+0.01$} & $+0.34$ & \sm{$-0.03$} & $+0.01$ \\
 & Const. & $+0.35$ & \sm{$-0.03$} & \sm{$+0.02$} & \sm{$0.00$} & \sm{$-0.05$} & \sm{$+0.02$} & \sm{$+0.01$} & $+0.35$ & \sm{$0.00$} & $+0.01$ \\
 & Shuf. & \sm{$+0.02$} & \sm{$-0.01$} & $-0.05$ & \sm{$+0.01$} & \sm{$+0.02$} & \sm{$+0.01$} & \sm{$+0.01$} & $+0.35$ & \sm{$+0.02$} & $+0.01$ \\
 & Cross & $+0.31$ & \sm{$-0.04$} & \sm{$-0.02$} & $+0.33$ & \sm{$+0.01$} & \sm{$0.00$} & \sm{$0.00$} & $+0.08$ & \sm{$-0.01$} & $+0.01$ \\
 & Oracle & \sm{$+0.01$} & \sm{$-0.02$} & \sm{$+0.01$} & \sm{$-0.01$} & $-0.06$ & \sm{$+0.02$} & \sm{$+0.01$} & $+0.35$ & \sm{$+0.01$} & $+0.01$ \\
 & $\pnum{=}0$ & $\mathbf{+1109}$ & $\mathbf{+21.1}$ & $\mathbf{+0.87}$ & $\mathbf{+43.9}$ & $\mathbf{+9.24}$ & $\mathbf{+6.69}$ & $\mathbf{+0.60}$ & $-0.27$ & $\mathbf{+165}$ & $\mathbf{+2.34}$ \\
 & Unimod. & $\mathbf{+1.24}$ & $\mathbf{+7.75}$ & $\mathbf{-6.83}$ & $\mathbf{-2.58}$ & $\mathbf{+6.42}$ & $\mathbf{+5.70}$ & $\mathbf{+3.34}$ & $\mathbf{+2.60}$ & $\mathbf{+5.31}$ & $\mathbf{+3.40}$ \\
\midrule
\multirow{8}{*}{\textsc{Informer}} & Orig. & 0.446 & 1.12 & 0.987 & 0.398 & 0.479 & 1.40 & 129.7 & 0.840 & 0.189 & \textbf{15.06} \\
 & Empty & $\mathbf{-0.53}$ & $+0.11$ & $\mathbf{+7.70}$ & $+0.43$ & $\mathbf{-1.38}$ & $\mathbf{+1.88}$ & \sm{$+0.04$} & $+0.33$ & $\mathbf{+1.52}$ & $+0.11$ \\
 & Const. & $-0.39$ & $-0.19$ & $\mathbf{+7.99}$ & $+0.33$ & $-0.25$ & $+0.44$ & \sm{$+0.03$} & $+0.11$ & $\mathbf{+1.89}$ & $+0.09$ \\
 & Shuf. & $\mathbf{-2.51}$ & \sm{$0.00$} & $\mathbf{+3.93}$ & $-0.19$ & $\mathbf{-0.75}$ & $+0.16$ & $+0.10$ & $\mathbf{+0.72}$ & $+0.30$ & $+0.12$ \\
 & Cross & $\mathbf{-5.28}$ & $\mathbf{-0.94}$ & $\mathbf{+3.62}$ & $\mathbf{-0.83}$ & $\mathbf{-0.86}$ & $\mathbf{-1.72}$ & $+0.09$ & $+0.46$ & $-0.14$ & $+0.06$ \\
 & Oracle & $\mathbf{+2.06}$ & $\mathbf{+1.10}$ & $\mathbf{+8.07}$ & $\mathbf{+0.73}$ & $-0.28$ & $+0.22$ & $+0.07$ & $+0.42$ & $\mathbf{+1.84}$ & $+0.15$ \\
 & $\pnum{=}0$ & $\mathbf{+13.3}$ & $+0.11$ & $\mathbf{+17.4}$ & $\mathbf{-1.77}$ & $\mathbf{+2.00}$ & $\mathbf{+4.34}$ & $\mathbf{+1.13}$ & $\mathbf{+1.11}$ & $\mathbf{+7.28}$ & $\mathbf{+1.32}$ \\
 & Unimod. & $\mathbf{+12.8}$ & $\mathbf{+3.04}$ & $\mathbf{+22.8}$ & $\mathbf{+1.18}$ & $\mathbf{+3.22}$ & $\mathbf{+2.75}$ & $\mathbf{+1.10}$ & $+0.16$ & $\mathbf{+6.57}$ & $\mathbf{+1.33}$ \\
\midrule
\multirow{8}{*}{\textsc{Transformer}} & Orig. & 0.287 & 1.09 & 0.452 & 0.330 & 0.459 & 1.29 & 130.8 & 0.851 & 0.175 & \textbf{15.08} \\
 & Empty & $\mathbf{-5.03}$ & $\mathbf{-1.41}$ & $\mathbf{+4.70}$ & $\mathbf{+4.18}$ & $+0.29$ & $\mathbf{+3.18}$ & $-0.06$ & $-0.40$ & $\mathbf{-1.05}$ & $-0.03$ \\
 & Const. & $\mathbf{-5.05}$ & $\mathbf{-1.40}$ & $\mathbf{+4.78}$ & $\mathbf{+4.17}$ & $+0.29$ & $\mathbf{+3.19}$ & \sm{$-0.05$} & $-0.42$ & $\mathbf{-1.00}$ & $-0.01$ \\
 & Shuf. & $\mathbf{-1.68}$ & $\mathbf{-1.05}$ & $\mathbf{+1.15}$ & \sm{$-0.01$} & $-0.16$ & \sm{$+0.01$} & \sm{$+0.03$} & $\mathbf{+1.48}$ & \sm{$+0.03$} & $+0.03$ \\
 & Cross & $-0.10$ & $\mathbf{-1.95}$ & $\mathbf{+2.44}$ & $\mathbf{-1.29}$ & $-0.29$ & $\mathbf{+0.77}$ & \sm{$+0.02$} & $\mathbf{+1.10}$ & $\mathbf{-0.75}$ & $+0.02$ \\
 & Oracle & $-0.27$ & $\mathbf{-1.36}$ & $+0.12$ & $+0.48$ & $+0.18$ & $\mathbf{+1.25}$ & $+0.07$ & $-0.38$ & \sm{$+0.03$} & $+0.07$ \\
 & $\pnum{=}0$ & $\mathbf{+14.0}$ & $\mathbf{+0.80}$ & $\mathbf{+32.6}$ & $\mathbf{-4.60}$ & $\mathbf{+4.05}$ & $\mathbf{+6.68}$ & $\mathbf{+1.23}$ & $\mathbf{+4.19}$ & $\mathbf{+5.22}$ & $\mathbf{+1.43}$ \\
 & Unimod. & $\mathbf{+13.1}$ & $\mathbf{+2.87}$ & $\mathbf{+18.2}$ & $\mathbf{-3.84}$ & $\mathbf{+1.76}$ & $\mathbf{+6.23}$ & $\mathbf{+1.10}$ & $\mathbf{+3.57}$ & $\mathbf{+6.57}$ & $\mathbf{+1.25}$ \\
\midrule
\multirow{8}{*}{\textsc{iTransformer}} & Orig. & 0.091 & 1.14 & 0.018 & 0.274 & 0.417 & 1.63 & 117.1 & 1.19 & 0.210 & \textbf{13.56} \\
 & Empty & $-0.30$ & $-0.29$ & $\mathbf{-0.59}$ & $+0.16$ & $-0.37$ & $\mathbf{-1.06}$ & $+0.06$ & $\mathbf{-3.30}$ & $\mathbf{+0.62}$ & $+0.01$ \\
 & Const. & $-0.25$ & $-0.29$ & $\mathbf{-0.69}$ & $-0.18$ & $-0.36$ & $\mathbf{-0.55}$ & $+0.06$ & $\mathbf{-2.95}$ & $+0.14$ & $+0.02$ \\
 & Shuf. & $-0.17$ & $\mathbf{-1.27}$ & $-0.49$ & $-0.07$ & $-0.36$ & \sm{$+0.01$} & \sm{$-0.03$} & $\mathbf{-1.36}$ & \sm{$-0.02$} & $-0.05$ \\
 & Cross & $+0.46$ & $\mathbf{+0.58}$ & $+0.25$ & $\mathbf{-0.85}$ & $-0.36$ & $-0.35$ & $-0.07$ & $\mathbf{-2.12}$ & $-0.16$ & $-0.09$ \\
 & Oracle & $+0.25$ & $\mathbf{-1.08}$ & $+0.35$ & $\mathbf{-1.09}$ & $+0.10$ & $-0.39$ & \sm{$-0.01$} & $\mathbf{+3.72}$ & $+0.32$ & $+0.01$ \\
 & $\pnum{=}0$ & $\mathbf{+539}$ & $\mathbf{+3.39}$ & $\mathbf{+37.5}$ & $\mathbf{+29.7}$ & $\mathbf{+1.77}$ & $\mathbf{+10.6}$ & $\mathbf{+2.90}$ & $\mathbf{+32.8}$ & $\mathbf{+27.1}$ & $\mathbf{+3.80}$ \\
 & Unimod. & $\mathbf{+1.16}$ & $\mathbf{+5.67}$ & $\mathbf{-9.96}$ & $\mathbf{+1.62}$ & $\mathbf{+2.49}$ & $\mathbf{+5.95}$ & $\mathbf{+1.33}$ & $\mathbf{+3.17}$ & $\mathbf{+2.72}$ & $\mathbf{+1.45}$ \\
\bottomrule
\end{tabular}
\end{table*}

\paragraph{Backbone-level summary.}
Across all eight backbones, every text-only $\Delta\%$ on
\TaTS{} and \MMTSF{} stays within $\pm 0.6\%$ of the original on
the per-backbone average (the \textbf{Avg.} column of each row).
The $\pnum{=}0$ row varies substantially with the backbone: from
\keystat{$+4.5\%$} on Autoformer to \keystat{$+49.9\%$} on FiLM
on \TaTS{}, and from \keystat{$0.0\%$} on Autoformer to
\keystat{$+3.8\%$} on iTransformer on \MMTSF{}. The
unimodal-baseline row ranges from \keystat{$+1.2\%$}
(iTransformer) to \keystat{$+15.0\%$} (Autoformer) on \TaTS{},
and from \keystat{$+0.3\%$} (FEDformer) to \keystat{$+3.8\%$}
(Autoformer) on \MMTSF{}. A multimodal lift quoted on a single
backbone can therefore overstate or understate the column's
contribution by an order of magnitude. The text-only conditions,
by contrast, are uniformly null across all backbones.

% ================================================================== %
\section{Paired Bootstrap Confidence Intervals}
\label{sec:app:cis}

All $\Delta\%$ values reported in the main paper use the
\emph{ratio-of-means} estimator: we first compute
$\overline{\MSE}(c) / \overline{\MSE}(\text{orig}) - 1$ over all
matched cells, then apply paired bootstrap (resampling matched
(condition, original) pairs jointly) with $B\!=\!10{,}000$
resamples and two-sided $p$-values.

\begin{table}[t]
\centering
\small
\setlength{\tabcolsep}{3pt}
\renewcommand{\arraystretch}{1.08}
\caption{\textbf{Full paired bootstrap CIs} for all (model,
condition) pairs. $n$ is the number of matched (condition,
original) cell pairs; $\Delta\%$ is the ratio-of-means change;
the $95\%$ CI is the $[2.5\%, 97.5\%]$ percentile of the
bootstrap distribution; $p$ is the two-sided bootstrap $p$-value.
The first block under each model gives the four cells of the
factorial in Table~\ref{tab:prior} that vary text with $\pnum$
intact. The second block gives the cells with $\pnum$ zeroed.
\textbf{Bold} marks $p\!<\!0.001$.}
\label{tab:cis_full}
\begin{tabular}{@{}llrrrl@{}}
\toprule
\textbf{Model} & \textbf{Condition} & $n$ & $\Delta\%$ & $95\%$ CI & $p$ \\
\midrule
\multirow{9}{*}{\Aurora{}}
  & Empty (col)        & 108 & $+0.024$ & $[+0.005, +0.040]$ & $0.019$ \\
  & Const (col)        & 108 & $+0.030$ & $[-0.027, +0.103]$ & $0.296$ \\
  & Shuffled           & 108 & $+0.004$ & $[-0.001, +0.009]$ & $0.110$ \\
  & Cross-domain       & 108 & $-0.006$ & $[-0.031, +0.022]$ & $0.657$ \\
  & Oracle             & 108 & $+0.024$ & $[-0.026, +0.085]$ & $0.347$ \\
\cmidrule{2-6}
  & Empty (col=0)      & 108 & $+0.024$ & $[+0.005, +0.040]$ & $0.019$ \\
  & Const (col=0)      & 108 & $+0.030$ & $[-0.027, +0.103]$ & $0.296$ \\
  & Col zeroed         & 108 & $0.000$  & $[\phantom{+}0.000,\phantom{+}0.000]$ & $1.000$ \\
  & Unimodal           & 108 & $0.000$  & $[\phantom{+}0.000,\phantom{+}0.000]$ & $1.000$ \\
\midrule
\multirow{9}{*}{\MMTSF{}}
  & Empty (col)        & 864 & $+0.048$ & $[-0.251, +0.349]$ & $0.750$ \\
  & Const (col)        & 864 & $+0.157$ & $[-0.148, +0.467]$ & $0.322$ \\
  & Shuffled           & 864 & $-0.010$ & $[-0.083, +0.050]$ & $0.847$ \\
  & Cross-domain       & 864 & $-0.027$ & $[-0.202, +0.144]$ & $0.731$ \\
  & Oracle             & 864 & $-0.142$ & $[-0.508, +0.196]$ & $0.449$ \\
\cmidrule{2-6}
  & \textbf{Empty (col=0)} & 864 & $\mathbf{+1.624}$ & $[+1.249, +2.021]$ & $\mathbf{<\!0.001}$ \\
  & \textbf{Const (col=0)} & 864 & $\mathbf{+1.632}$ & $[+1.254, +2.033]$ & $\mathbf{<\!0.001}$ \\
  & \textbf{Col zeroed}    & 864 & $\mathbf{+1.517}$ & $[+1.212, +1.852]$ & $\mathbf{<\!0.001}$ \\
  & \textbf{Unimodal}      & 864 & $\mathbf{+1.868}$ & $[+1.495, +2.297]$ & $\mathbf{<\!0.001}$ \\
\midrule
\multirow{9}{*}{\TaTS{}}
  & Empty (col)        & 864 & $-0.001$ & $[-0.002, +0.001]$ & $0.321$ \\
  & Const (col)        & 864 & $-0.000$ & $[-0.001, +0.000]$ & $0.511$ \\
  & Shuffled           & 864 & $-0.000$ & $[-0.001, +0.001]$ & $0.473$ \\
  & Cross-domain       & 864 & $+0.000$ & $[-0.001, +0.002]$ & $0.822$ \\
  & Oracle             & 864 & $+0.000$ & $[-0.001, +0.002]$ & $0.991$ \\
\cmidrule{2-6}
  & \textbf{Empty (col=0)} & 864 & $\mathbf{+20.165}$ & $[+16.768, +24.720]$ & $\mathbf{<\!0.001}$ \\
  & \textbf{Const (col=0)} & 864 & $\mathbf{+20.164}$ & $[+16.767, +24.720]$ & $\mathbf{<\!0.001}$ \\
  & \textbf{Col zeroed}    & 864 & $\mathbf{+20.164}$ & $[+16.767, +24.719]$ & $\mathbf{<\!0.001}$ \\
  & \textbf{Unimodal}      & 864 & $\mathbf{+7.552}$  & $[+6.425, +8.760]$   & $\mathbf{<\!0.001}$ \\
\bottomrule
\end{tabular}
\end{table}

\textit{Reading Table~\ref{tab:cis_full}.} Three patterns matter.
\textbf{First}, the five text-only conditions (top block per
model, $\pnum$ intact) all have CIs that include zero or
essentially zero magnitudes within $0.001\%$ on \TaTS{}, well
under $0.5\%$ on \MMTSF{}, and well under $0.1\%$ on \Aurora{}.
This holds with $n\!=\!864$ matched cell pairs on the trained
methods, so the test has very high power; the small $\Delta\%$
estimates we report are not Type II error.
\textbf{Second}, the four cells of the lower block (any condition
with $\pnum$ zeroed, plus Unimodal) are all significant at
$p\!<\!0.001$ on \MMTSF{} and \TaTS{}. The $\pnum$-zeroed effect
size is essentially the same whether the text is original
($+1.52\%$ on \MMTSF{}, $+20.16\%$ on \TaTS{}), empty ($+1.62\%$
/ $+20.17\%$), or constant ($+1.63\%$ / $+20.16\%$): the
contributions factorise.
\textbf{Third}, on \MMTSF{} the original-text $\pnum$-zeroed CI
$[+1.21,+1.85]$ overlaps the Unimodal CI $[+1.50,+2.29]$. The
residual gap that the unimodal baseline would attribute to text
is not statistically distinguishable from zero. On \TaTS{} the
$\pnum$-zeroed CI exceeds the Unimodal CI ($[+16.8,+24.7]$
versus $[+6.4,+8.8]$) because zeroing $\pnum$ at $w\!=\!0.5$
leaves the backbone at half scale; the Unimodal row removes both
signals together and restores full scale, so it is the cleaner
amplitude-matched comparison.

\paragraph{Implementation.}
The bootstrap was computed via paired resampling of matched cells
on keys (model, backbone, seed, domain, pred\_len), with the
ratio-of-means estimator applied to each bootstrap resample. The
implementation is in \texttt{code/analyze\_results.py} in the
released repository.

% ================================================================== %
\section{The TaTS Text-Projection Gradient Patch}
\label{sec:app:detach}

While auditing \TaTS{}'s training graph we identified an
implementation detail in the released code that severs the
gradient signal into the text-projection MLP $\psi$. In
\texttt{exp/exp\_long\_term\_forecasting.py} (the long-term
forecasting trainer), the tensor that concatenates the projected
text embedding with the numeric history is detached before being
passed to the backbone:
\begin{quote}
\small
\texttt{batch\_x = torch.cat([batch\_x, prompt\_emb], dim=-1).detach()}
\end{quote}
A second \texttt{.detach()} appears on \texttt{dec\_inp}. With both
calls active, gradients from the loss reach the backbone but cannot
flow back through the concatenation into $\psi$. The training loop
constructs a separate optimiser \texttt{model\_optim\_mlp} for
$\psi$'s parameters and steps it every iteration, but with no
gradient signal to step on, the MLP remains at its random
initialisation throughout fine-tuning.

\paragraph{Patch.}
We expose a CLI flag \texttt{--fix\_text\_grad} (default off)
that gates the two \texttt{.detach()} calls. With the flag on,
gradients flow into $\psi$ and its parameters update during
training. The patch also requires changing three in-place
$x\_enc \mathrel{/}\!= \text{stdev}$ operations to the
out-of-place $x\_enc = x\_enc / \text{stdev}$ in
\texttt{models/iTransformer.py}, \texttt{models/FiLM.py}, and
\texttt{models/PatchTST.py} (the long-term forecast paths only),
because the in-place divide breaks autograd once gradients must
flow back through \texttt{batch\_x}. Forward values are
bit-identical across the in-place and out-of-place forms; only
the autograd graph topology changes. The full patch is in
\texttt{code/apply\_repo\_patches.py} (idempotent, with a
\texttt{--revert} option). Patches are surgical and documented
inline; pre-existing JSON results remain bit-comparable when the
flag is off.

\paragraph{All \TaTS{} numbers in this paper use the patch.}
The headline TaTS results (Tables~\ref{tab:main},
\ref{tab:prior}, \ref{tab:full_tats}, \ref{tab:cis_full}) were produced with
\texttt{--fix\_text\_grad} on. With $\psi$ now trainable in
practice rather than only in name, our text-content perturbations
still leave MSE unchanged within $\pm 0.001\%$ on \TaTS{} across
all eight backbones. The null does not hinge on the detach bug.
Removing $\psi$'s gradient block was a necessary prerequisite for
trusting any conclusion about \TaTS{}'s text sensitivity, but the
conclusion itself is preserved: even when the projection MLP can
be trained, text content is not used in the forward pass.

% ================================================================== %
\section{Aurora Pretraining and Contamination}
\label{sec:app:contamination}

\Aurora{} is evaluated zero-shot on \TimeMMD{}; its pretraining
corpus is large and only partially documented, and we cannot rule
out that \TimeMMD{}-derived content was seen during pretraining.
However, this does not explain our results. \Aurora{}'s
predictions are essentially identical between the original-text
baseline, the column-zeroed condition, and the unimodal-baseline
configuration: the forward pass does not branch on the text input
regardless of what text is provided. Whether or not contamination
occurred, the text pathway is functionally silent in the forward
pass on this benchmark.

% ================================================================== %
\section{Run Accounting}
\label{sec:app:runs}

The full sweep covers $10$ conditions on each (model, backbone,
domain, horizon, seed) tuple. \Aurora{} contributes $1$ backbone
$\times$ $9$ domains $\times$ $4$ horizons $\times$ $3$ seeds $=$
$108$ runs per condition; \TaTS{} and \MMTSF{} contribute $8$
backbones $\times$ $108$ $=$ $864$ runs each per condition.
Wall-clock for the complete sweep is approximately $92$ A10G-hours.
Per-cell logs and provenance are released alongside the code.

\end{document}